\documentclass[sigconf]{acmart}

\setcopyright{none}
\renewcommand\footnotetextcopyrightpermission[1]{}
\AtBeginDocument{%
  \providecommand\BibTeX{{%
    \normalfont B\kern-0.5em{\scshape i\kern-0.25em b}\kern-0.8em\TeX}}}

\usepackage{graphicx}	% required for `\includegraphics' (yatex added)
\usepackage{manyfoot}
\usepackage{algorithm}
\usepackage{algorithmic}
\usepackage[caption=false,font=footnotesize]{subfig}

\begin{document}

\title{CSI2Image: Image Reconstruction from Channel State Information Using Generative Adversarial Networks}

\author{Sorachi Kato}
\email{kato.sorachi@ist.osaka-u.ac.jp}
\affiliation{%
  \institution{
  Division of Electronic and Information Engineering,\\
  School of Engineering,\\
  Osaka University}
  \city{Suita}
  \state{Osaka}
  \postcode{565--0871}
  \country{Japan}}

\author{Takeru Fukushima}
\author{Tomoki Murakami}
\author{Hirantha Abeysekera}
\affiliation{%
\institution{Access Network Service Systems Laboratories,\\
NTT Corporation,}
\city{Yokosuka}
\state{Kanagawa}
\postcode{239--0847}
  \country{Japan}}

\author{Yusuke Iwasaki}
\author{Takuya Fujihashi}
\author{Takashi Watanabe}
\affiliation{%
  \institution{
  Graduate School of Information Science and Technology,\\
  Osaka University}
  \city{Suita}
  \state{Osaka}
  \postcode{565--0871}
  \country{Japan}}

\author{Shunsuke Saruwatari}
\email{saru@ist.osaka-u.ac.jp}
\affiliation{%
\institution{
Graduate School of Information Science and Technology,\\
Osaka University}
\city{Suita}
\state{Osaka}
\postcode{565--0871}
\country{Japan}}

\renewcommand{\shortauthors}{Kato and Fukushima, et al.}

\begin{abstract}
This study aims to find the upper limit of the wireless sensing capability of acquiring physical space information.
This is a challenging objective, because at present, wireless sensing studies continue to succeed in acquiring novel phenomena.
Thus, although a complete answer cannot be obtained yet, a step is taken towards it here.
To achieve this, CSI2Image, a novel channel-state-information (CSI)-to-image conversion method based on generative adversarial networks (GANs), is proposed.
The type of physical information acquired using wireless sensing can be estimated by checking wheth\-er the reconstructed image captures the desired physical space information.
Three types of learning methods are demonstrated: gen\-er\-a\-tor-only learning, GAN-only learning, and hybrid learning.
Evaluating the performance of CSI2Image is difficult, because both the clarity of the image and the presence of the desired physical space information must be evaluated.
To solve this problem, a quantitative evaluation methodology using an object detection library is also proposed.
CSI2Image was implemented using IEEE 802.11ac compressed CSI, and the evaluation results show that the image was successfully reconstructed.
The results demonstrate that gen\-er\-a\-tor-only learning is sufficient for simple wireless sensing problems, but in complex wireless sensing problems, GANs are important for reconstructing generalized images with more accurate physical space information.
\end{abstract}

\begin{CCSXML}
<ccs2012>
   <concept>
       <concept_id>10010520.10010553</concept_id>
       <concept_desc>Computer systems organization~Embedded and cy\-ber-phys\-i\-cal systems</concept_desc>
       <concept_significance>100</concept_significance>
       </concept>
   <concept>
       <concept_id>10010520.10010553.10003238</concept_id>
       <concept_desc>Computer systems organization~Sensor networks</concept_desc>
       <concept_significance>100</concept_significance>
       </concept>
 </ccs2012>
\end{CCSXML}



\maketitle

\section{Introduction}

This study considers the upper limit of the wireless sensing capability of acquiring physical space information. Wireless sensing enables us to obtain a variety of data in physical space by only deploying access points (APs).
Several studies have already shown the possibility of extracting physical space information from radio waves.
In particular, channel state information (CSI)--based methods are improving the practical feasibility of wireless sensing. This is because CSI, which is used for multiple-input multiple-output (MIMO) communication, is easily acquired from commercial Wi-Fi devices.
Using Wi-Fi CSI, state-of-the-art studies have already achieved remarkable results.
In the future, Wi-Fi may become a sensing platform; the IEEE 802.11 wireless LAN working group has established a study group for WLAN sensing.
The details of wireless sensing are discussed in Section \ref{sec:wireless-sensing}.

To understand the upper limit of the wireless sensing capability of acquiring physical space information, this study attempts to reconstruct images from CSI obtained from off-the-shelf Wi-Fi devices. 
If the conversion from CSI to images corresponding to the physical space can be realized, the possibly of extracting physical space information using CSI can be approximately estimated.
In addition, because the eye is the most high-resolution sensor in the human body, the images serve as human-understandable information.
Furthermore, object detection technology, which has developed in conjunction with the emergence of deep learning and the next generation of applications such as automated driving, can be used to automatically build learning data without manual labeling.

\begin{figure}[b]
  \begin{center}
   \subfloat[Learning phase]{\label{fig:learning_csi_image_recognition}\includegraphics[scale=0.29]{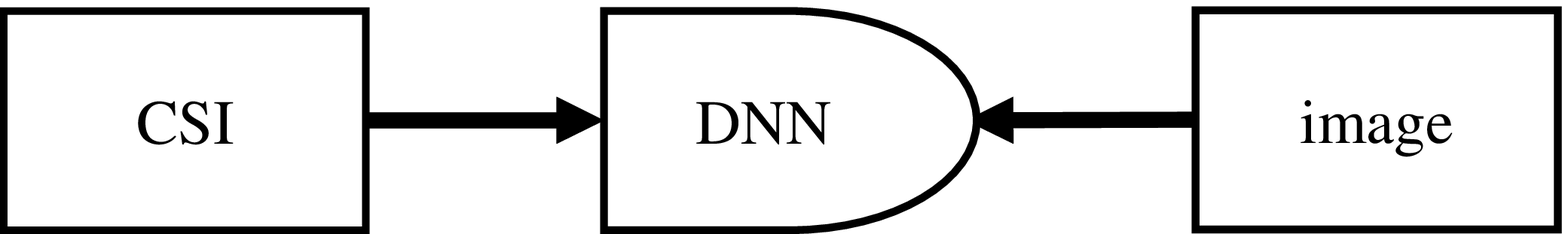}}
   \hfill
   \subfloat[Recognition phase]{\label{fig:exec_csi_recognition}\includegraphics[scale=0.29]{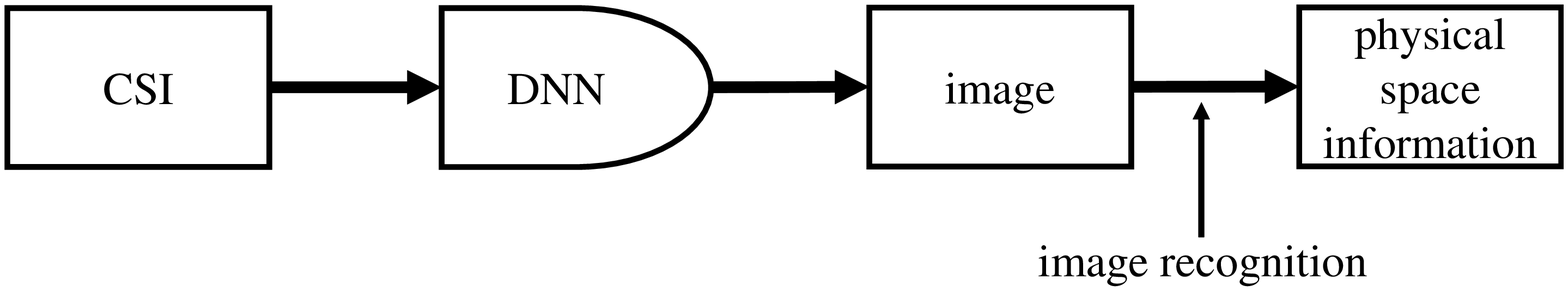}}
   \hfill
   \subfloat[Our demonstration]{\label{fig:example_csi_recognition}\includegraphics[scale=0.29]{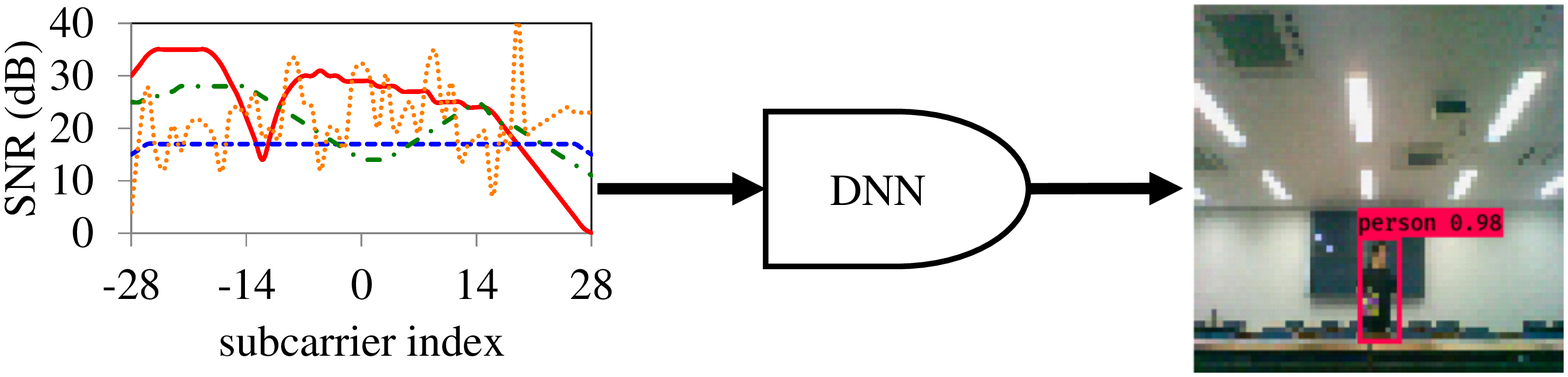}}
  \caption{Application example of CSI2Image with object detection}
  \label{fig:csi_image_recognition}
  \end{center}
\end{figure}

Figures~\ref{fig:csi_image_recognition}  shows an application example of CSI-to-image conversion: automatic wireless sensing model generation.
The generation consists of two phases: the learning phase and the recognition phase. 
In the learning phase shown in Figure~\ref{fig:csi_image_recognition}\subref{fig:learning_csi_image_recognition}, the system simultaneously captures the CSI and images of the target space, following which the system trains a deep neural network (DNN) with the captured CSI and the images.
Finally, the system extracts the physical space information from the image reconstructed from the captured CSI using the trained DNN, as shown in Figure~\ref{fig:csi_image_recognition}\subref{fig:exec_csi_recognition}.
Figure~\ref{fig:csi_image_recognition}\subref{fig:example_csi_recognition} shows a practical example of the automatic wireless sensing model generation; this is demonstrated as an evaluation in Section \ref{sec:evaluation}.

Considering this, this paper proposes CSI2Image, a novel wireless sensing method to convert radio information into images corresponding to the target space using DNN.
To the best of our knowledge, this is the first time CSI-to-image conversion has been a\-chieved using GANs.
From the perspective of CSI-to-image conversion without GANs, a few related studies have been conducted \cite{kefayati20:ieee, ieee19guo}.
Wi2Vi \cite{kefayati20:ieee} uses the video recovered by CSI when a security camera is disabled due to power failure, malfunction, or attack.
Under normal conditions, Wi2Vi extracts the background image from the camera image, detects a person using the difference between the background image and the image, and learns by associating it with the CSI.
Under critical conditions, Wi2Vi generates an image by superimposing the detected user onto the background image.
\cite{ieee19guo} has successfully generated pose images generated from skeletons from CSI by learning the relationship between the skeleton model of human posture and CSI.
\cite{kefayati20:ieee, ieee19guo} are application-specific approaches, using application-specific information such as background images and skeleton models. In contrast, the present study focuses on a general-purpose CSI-to-image conversion method using GANs.

The main contributions of this paper are as follows:
\begin{itemize}
 \item The use of GANs for CSI-to-image conversion is proposed, implemented, and evaluated.
       In particular, because simply introducing GANs is insufficient, this paper shows three methods of learning the conversion model: generator-only learning, GAN-only learning, and hybrid learning.
 \item Novel position-detection-based quantitative evaluation meth\-od\-ol\-ogy to evaluate the performance of CSI-to-image conversion is demonstrated. 
       Specifically, Section \ref{sec:evaluation} quantitatively shows that the use of GANs enables the successful reconstruction of more generalized images from CSI compared to generator-only learning.
 \item Empirical evaluation using off-the-shelf devices is performed using compressed CSI, which can be acquired from IEEE 802.11ac devices.
       The obtained results can be easily reproduced using an off-the-shelf USB camera, a Raspberry Pi, and a packet capture tool.
 \end{itemize}

The remainder of this paper is organized as follows.
Section 2 describes related works on wireless sensing and GANs.
Section 3 proposes CSI2Image with three generator learning structures: generator-only learning, GAN-only learning, and hybrid learning.
Section 4 presents the qualitative and quantitative evaluation of the three learning structures proposed in Section 3; the quantitative evaluation methodology is also proposal for the evaluation of CSI-to-image conversion.
Finally, conclusions are presented in Section 5.

\section{Related Works}

The present work explores the areas of wireless sensing and GANs.

\subsection{Wireless sensing}
\label{sec:wireless-sensing}

Several studies have already shown the possibility of extracting physical space information from radio waves using wireless sensing. This has been applied for various purposes, including device localization \cite{chang08:sensys,zhong09:sensys,xi10:sensys,dil11:sensys,xiong15:mobicom,he15:sensys,kotaru15:sigcomm,shangguan15:nsdi,vasisht16:nsdi,ma16:mobicom,khaledi17:mobicom,ma17:mobicom,chen17:conext,pefkianakis18:sensys,nandakumar18:sensys,soltanaghaei18:mobisys,bernhard18:sensys,xu19:sensys,tai19:mobisys}, 
device-free user localization \cite{adib15:nsdi,ohara15:ubicomp,shangguan15:nsdi,wang16:mobicom,fukushima19:vtc,xie19:mobicom,chen19:sensys},
gesture recognition \cite{pu13:mobicom,zhang18:mobicom,chi18:sensys,abdelnasser19:ieee},
device-free motion tracking \cite{joshi15:nsdi,li15:mobicom,zhu17:sensys}, 
RF imaging \cite{zhunge11:ieee,huang14:sensys,adib15:acm},
crowdedness estimation \cite{matsumoto19:vtc}, 
activity recognition \cite{wang15:mobicom,jiang18:mobicom}, 
hidden electronics detection \cite{li18:sensys}, 
respiratory monitoring \cite{adib15:chi, hillyard18:mobicom},
heart rate monitoring \cite{adib15:chi}, 
material sensing \cite{zhang19:mobicom,xie19:sensys}, 
soil sensing \cite{ding19:mobicom},
keystroke recognition \cite{ali15:mobicom}, 
emotion recognition \cite{zhao18:cacm}, 
human dynamics monitoring~\cite{guo17:sensys}, 
in-body device localization \cite{vasisht18:sigcomm}, 
object state change detection \cite{ohara17:imwut},
touch sensing \cite{gao18:nsdi},
device proximity detection \cite{pierson19:mobicom},
device orientation tracking \cite{wei16:mobicom}, and 
human detection through walls \cite{adib13:sigcomm,adib15:acm,yang15:mobicom}.
While these studies have explored new possibilities using an application-specific approach, the present work is unique in that it attempts to construct a general-purpose wireless sensing technique.

In terms of the physical layer, the proof of concept has been demonstrated in wireless communication devices such as specially customized hardware \cite{chang08:sensys,pu13:mobicom,adib13:sigcomm,adib14:nsdi,huang14:sensys,adib15:nsdi,adib15:chi,adib15:acm,zhao18:cacm,vasisht18:sigcomm,gao18:nsdi,nandakumar18:sensys}, 
mmWave \cite{pefkianakis18:sensys,li18:sensys}, 
UWB \cite{zhunge11:ieee,bernhard18:sensys},
RFID \cite{wang14:sigcomm,shangguan15:nsdi,yang15:mobicom,ma16:mobicom,wei16:mobicom,ma17:mobicom,luo19:nsdi,xu19:sensys,xiao19:imwut,xie19:sensys}, 
LoRa \cite{chen19:sensys}, 
IEEE 802.15.4 \cite{zhong09:sensys,xi10:sensys,dil11:sensys,chi18:sensys,matsumoto19:vtc},
Bluetooth \cite{chen17:conext,ayyalasomayajula18:conext}, 
IEEE 802.\-11n \cite{kotaru15:sigcomm,wang15:mobicom,ohara15:ubicomp,ali15:mobicom,wang16:mobicom,kotaru17:cvpr,zhu17:sensys,guo17:sensys,jiang18:mobicom,zhang18:mobicom,zhang19:mobicom,ding19:mobicom}, 
and IEEE 802.11ac \cite{murakami18:mdpi,fukushima19:vtc}.
The use of special customized hardware such as USRP \cite{usrp} and WARP \cite{khattab08:acm} enables the extraction of more detailed physical space information.
However, the use of commercially available equipment such as IEEE 802.11n, RFID, Bluetooth, and IEEE 802.15.4 is advantageous for deployment and the reproducibility of research results.
In particular, the emergence of CSI tools \cite{halperin11:sigcomm,xie15:mobicom,csitool,atherostool}has been particularly significant for the wireless sensing research community.
Commercially available IEEE 802.11n devices have been used not only to produce various research results, but they have also opened up possibilities for the deployment of wireless sensing.
However, at present, research using IEEE 802.11n faces the problem that only one section of IEEE 802.11n devices, Intel 5300 NIC, Atheros AR9390, AR9580, AR9590, AR9344, or QCA9558, can obtain CSI.
The present study uses the IEEE 802.11ac \cite{book13perahia,ieee16:ieeestd} compressed CSI.
The IEEE 802.11ac compressed CSI is standardized to reduce the overhead of the CSI feedback.
Compressed CSI can be acquired from any device that supports IEEE 802.11ac or IEEE 802.11ax.

\subsection{Generative adversarial networks}

GANs enable the generation of new data with the same statistics as the training data using a generative model \cite{goodfellow14:nips}, and they have been used in several applications \cite{wang19:arxiv,pan19:ieee}.
The generative model is constructed by alternately learning a generator and a discriminator in order to trick the discriminator.
This section introduces deep convolutional GAN (DCGAN) \cite{radford15:arxiv} and super-resolution GAN (SRGAN) \cite{ledig17:cvpr}, both of which are highly relevant to this study.

\begin{figure}[bt]
 \center
 \includegraphics[scale=0.25]{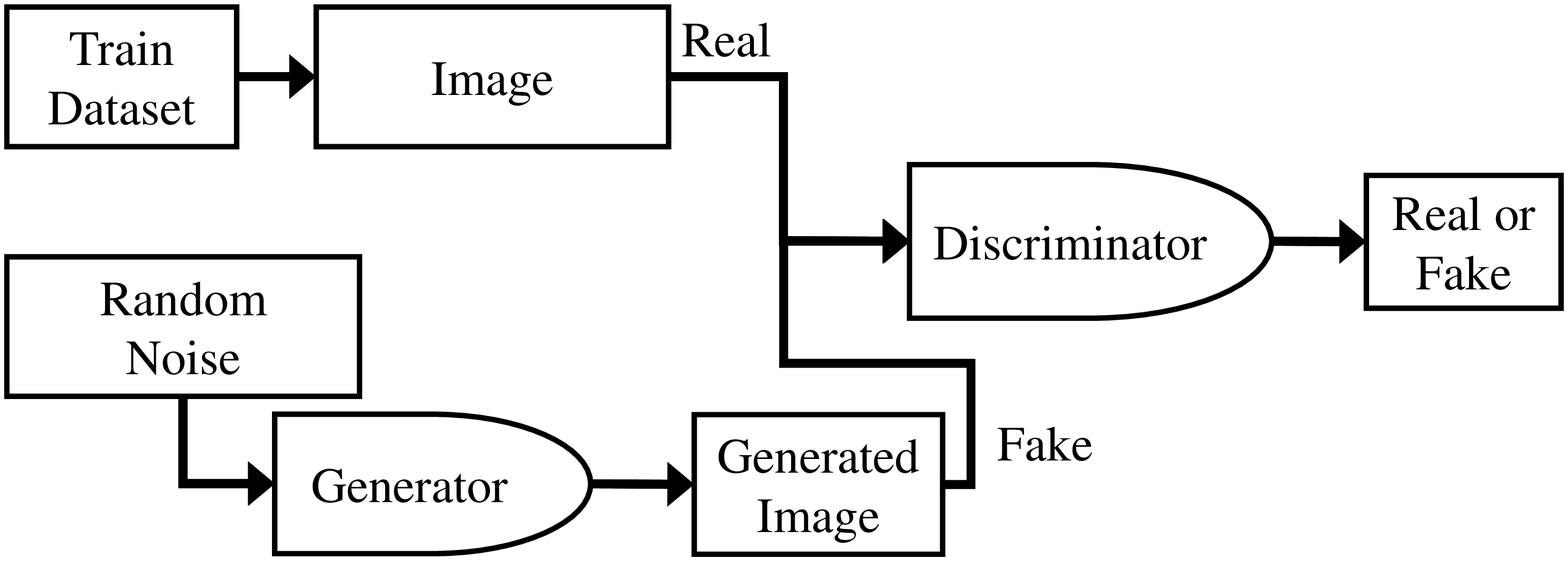}
 \caption{DCGAN}
 \label{fig:dcgan}
\end{figure}

DCGAN constructs a generative model to generate realistic fake images from random noise \cite{radford15:arxiv}.
Figure~\ref{fig:dcgan} shows the model structure of DCGAN.
DCGAN trains the discriminator to identify an image as real when the image is from the training dataset, and as fake when it is generated from random noise by the generator.
At the same time, DCGAN trains the generator to generate images (from random noise) that the discriminator identifies as real.
The generator is implemented using deep convolutional neural networks \cite{krizhevsky12:nips}.
As the generator and the discriminator learn to compete with each other, the generator is able to generate high-quality fake images.

\begin{figure}[bt]
 \center
 \includegraphics[scale=0.25]{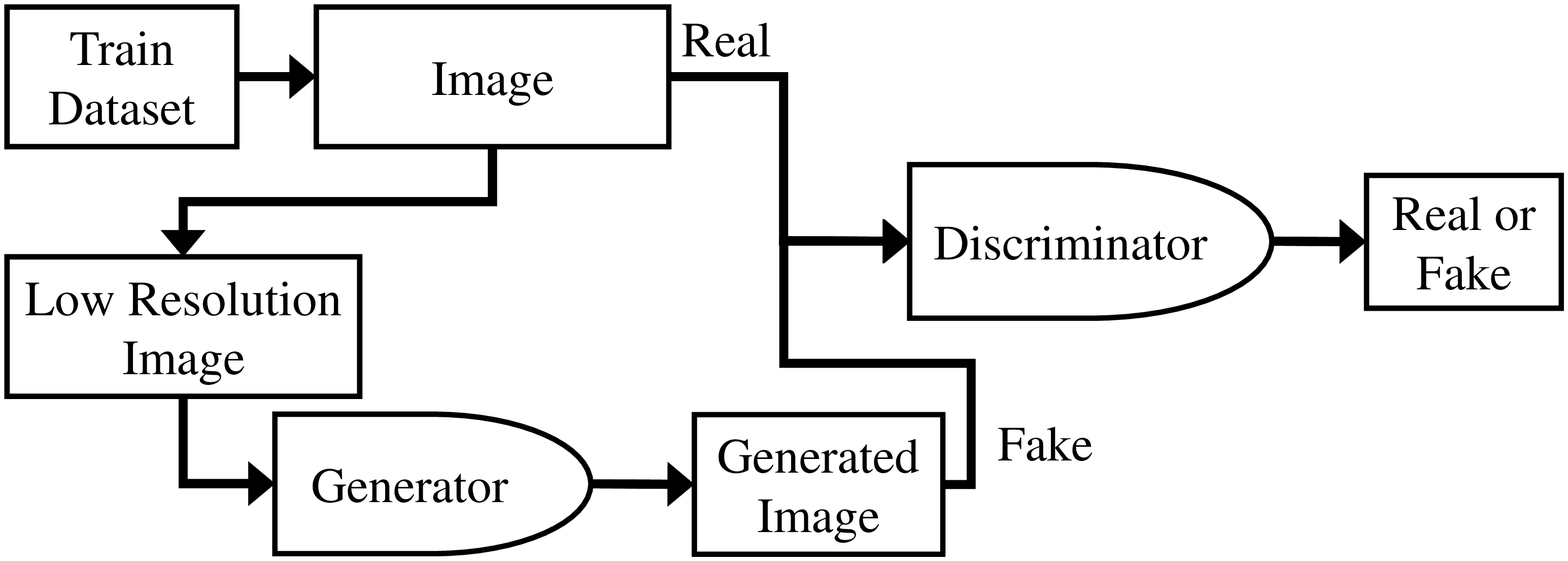}
 \caption{SRGAN}
 \label{fig:srgan}
\end{figure}

SRGAN generates high-resolution images from the corresponding low-resolution images \cite{ledig17:cvpr}.
Figure \ref{fig:srgan} shows the model structure of SRGAN.
SRGAN trains the discriminator to identify an image as real when it is from the training dataset, and as fake when it is generated from a low-resolution image by the generator.
At the same time, SRGAN trains the generator to attempt to generate images (from low-resolution images) that will be identified as real by the discriminator.
From DCGAN and SRGAN, it can be said that GANs can be used to create fake data that appears real or to recover real data from small amounts of data.

\section{CSI2Image: Image reconstruction from CSI}

\begin{figure}[bt]
 \center
 \includegraphics[scale=0.25]{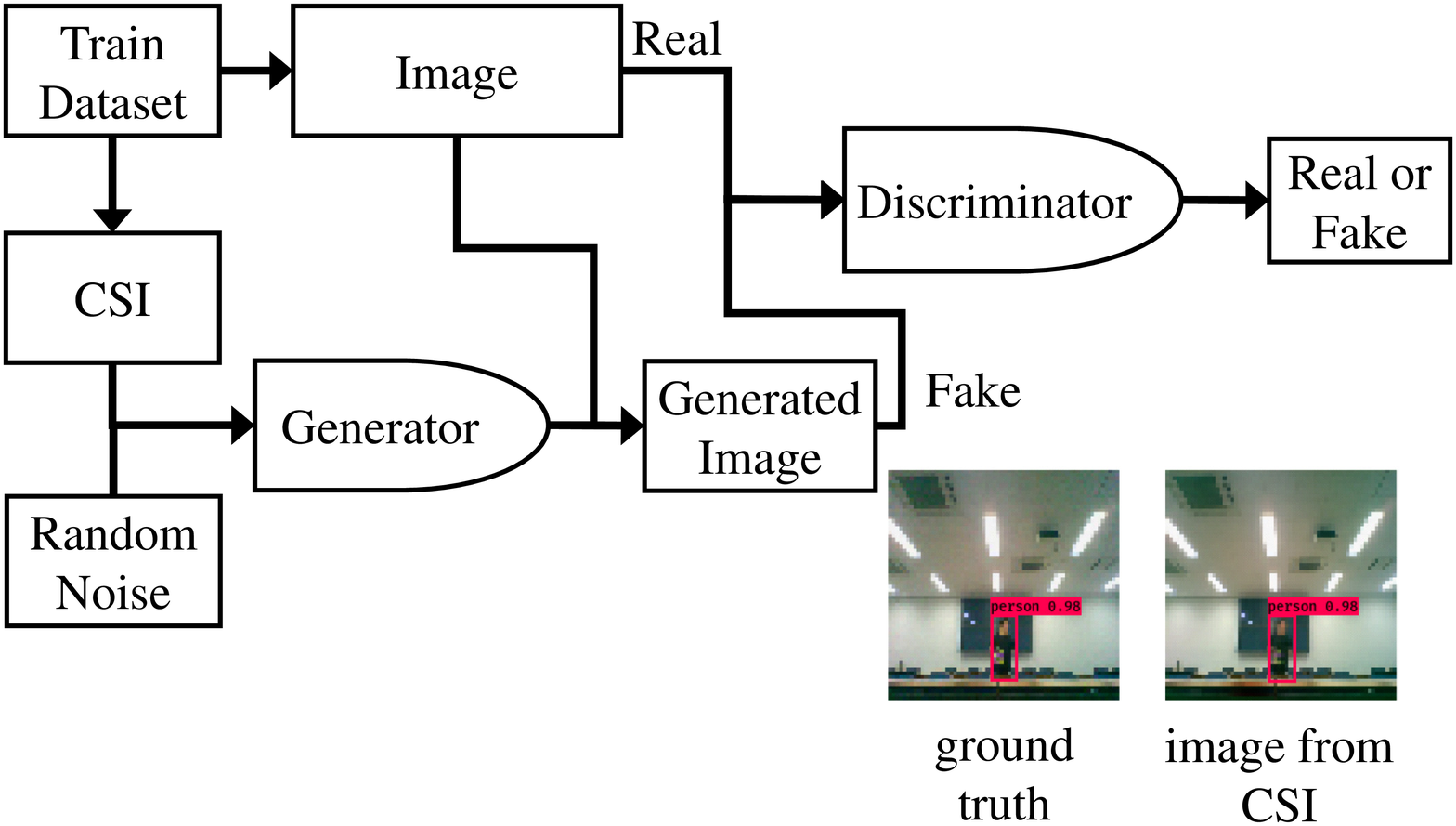}
 \caption{Overview}
 \label{fig:overview}
\end{figure}

Figure \ref{fig:overview} shows the entire system of the proposed CSI2\-Image.
CSI2\-Image is composed of training data, a generator, and a discriminator.
Section \ref{sec:training-data} shows the details of the training data, Section \ref{sec:generator-model} shows the model structure of the generator, and Section \ref{sec:discriminator-model} shows the model structure of the discriminator.
Note that this paper proposes three types of generator learning methods: generator-only learning, GAN-only learning, and hybrid learning.
Generator-only learning does not use a discriminator, as described in Section \ref{sec:learning-phase}.

\subsection{Training data}
\label{sec:training-data}

The training data of CSI2Image consist of simultaneously captured images and CSIs.
Full-color 64 $\times$ 64 pixel images and compressed CSI of $312$ dimensions, acquired with $2$ TX antennas, $3$ RX antennas, and $52$ subcarriers, are used.
The compressed CSI is used in off-the-shelf APs, smartphones, and PCs for their wireless communications, and the common format of CSI feedback is as specified in the IEEE 802.11ac standard \cite{book13perahia,ieee16:ieeestd}.

CSI2Image recovers the right singular matrix $\bf V$ from compressed CSI and uses the first column of the $\bf V$ as input data.
Note that the singular value decomposition of the CSI is expressed as follows.
    \begin{equation}
       \label{equ:svd}
       {\bf CSI} = {\bf U}{\bf S}{\bf V}^H \nonumber
    \end{equation}
where ${\bf U}$ is a left singular matrix, ${\bf S}$ is a diagonal matrix with singular values of CSI, and ${\bf V}$ is a right singular matrix.

The compressed CSI in IEEE 802.11ac includes the angle information $\phi$ and $\psi$.
The $\bf V$ is calculated with $\phi$ and $\psi$ by the following Equation (\ref{equ:givens}).
    \begin{equation}
       \label{equ:givens}
       {\bf V} = \left\{\prod_{k=1}^{{\rm min}(N, M-1)} \left[{\bf D}_k \prod_{l=k+1}^{M} {\bf G}_{l,k}^{T}(\psi_{k,k})\right]\right\}\widetilde{{\bf I}}_{M\times N},
    \end{equation}
where $M$ is the number of RX antennas, $N$ is the number of TX antennas, and $\widetilde{{\bf I}}_{M\times N}$ is the identity matrix in which zeros are inserted in the missing element if $N\neq M$.
${\bf D}_k$ is a diagonal matrix, expressed as follows:
\begin{equation}
 \label{mat:dk}
  {\bf D}_k=
  \begin{pmatrix}
   {\bf I}_{k-1} & 0 & 0 & \cdots & 0\\
   0 & e^{j\phi_{k,k}} & 0 & \cdots & 0\\
   0 & 0 & \ddots & 0 & 0 \\
   \vdots & \vdots & 0 & e^{j\phi_{M-1, k}} & 0 \\
   0 & 0 & 0 & 0 & 1 \\
  \end{pmatrix} \nonumber
\end{equation}
where ${\bf I}_{k-1}$ is $(k-1)\times(k-1)$ identity matrix.
${\bf G}_{l,k}(\psi)$ is the Givens rotation matrix
\begin{equation}
       \label{mat:g}
       {\bf G}_{l,k}=
       \scalebox{0.9}{$\displaystyle
       \begin{pmatrix}
       {\bf I}_{k-1} & 0 & 0 & 0 & 0\\
       0 & {\rm cos}(\psi) & 0 & {\rm sin}(\psi) & 0\\
       0 & 0 & {\bf I}_{l-k-1} & 0 & 0 \\
       0 & -{\rm sin}(\psi) & 0 & {\rm cos}(\psi) & 0 \\
       0 & 0 & 0 & 0 & {\bf I}_{M-1} \\
       \end{pmatrix}
       $} \nonumber
       \end{equation}
where ${\bf I}_{k-1}$ is $(k-1)\times(k-1)$ identity matrix.

\subsection{Generator network}
\label{sec:generator-model}

Table~\ref{table:gen} shows the network structure of the generator.
The compressed CSI of $312$ dimensions is input to the dense layer of 65,536 neurons with the Rectified Linear Unit (ReLU) layer, and the neurons are reshaped into a $8 \times 8 \times 1024$ tensor.
The tensor goes into the upsample layer, convolution layer with $3 \times 3$ kernel, batch normalization layer, and ReLU layer thrice.
Finally, it is also input to the convolution layer with the $3 \times 3$ kernel and activation function of tanh to obtain the output of $64 \times 64 \times 3$ tensor.
Adam is utilized as the optimizer of the generator network, whose learning rate is 0.0002, and momentum term is 0.5. 
The loss function for the generator network is the mean squared error (MSE)~\cite{ieeespm09zhou}.

\begin{table}[]
 \small
 \center
  \caption{Generator network}
  \label{table:gen}
  \begin{tabular}{l}
  \hline
  Input(312)                                                    \\
  Dense(65536)                                                  \\
  Activation("ReLU")                                            \\
  Reshape((8, 8, 1024))                                         \\
  UpSampling((2, 2))                                            \\
  Convolution(filters=512, kernel\_size=(3, 3), padding="same") \\
  BatchNormalization(momentum=0.8)                              \\
  Activation("ReLU")                                            \\
  UpSampling((2, 2))                                            \\
  Convolution(filters=256, kernel\_size=(3, 3), padding="same") \\
  BatchNormalization(momentum=0.8)                              \\
  Activation("ReLU")                                            \\
  UpSampling((2, 2))                                            \\
  Convolution(filters=128, kernel\_size=(3, 3), padding="same") \\
  BatchNormalization(momentum=0.8)                              \\
  Activation("ReLU")                                            \\
  Convolution(filters=64, kernel\_size=(3, 3), padding="same")  \\
  Activation("tanh")                                            \\
  \hline
  \end{tabular}
\end{table}

\subsection{Discriminator network}
\label{sec:discriminator-model}

Table \ref{table:dis} shows the network structure of the discriminator.
The input is a full-color image of $64 \times 64$ pixels.
The color image is then fed into four sets of the convolution layer, of a $3 \times 3$-size kernel with stride 2, batch normalization, LeakyReLU function ($\alpha$ = 0.2), and dropout of 0.25.
The output is then flattened and activated by a sigmoid function.
The output value is the range of 0 to 1.
The discriminator network uses the Adam optimizer, whose initial setting is the same as that of the generator network, and the loss function of binary cross-entropy.

\begin{table}[]
\center
\small
  \caption{discriminator network}
  \label{table:dis}
  \begin{tabular}{l}
  \hline
  Input((64, 64 , 3))                                                      \\
  Convolution(filters=32, kernel\_size=(3, 3), strides=2, padding="same")  \\
  Activation(``LeakyReLU'', alpha=0.2)                                       \\
  Dropout(0.25)                                                            \\
  Convolution(filters=64, kernel\_size=(3, 3), strides=2, padding="same")  \\
  BatchNormalization(momentum=0.8)                                         \\
  Activation("LeakyReLU", alpha=0.2)                                       \\
  Dropout(0.25)                                                            \\
  Convolution(filters=128, kernel\_size=(3, 3), strides=2, padding="same") \\
  BatchNormalization(momentum=0.8)                                         \\
  Activation("LeakyReLU", alpha=0.2)                                       \\
  Dropout(0.25)                                                            \\
  Convolution(filters=256, kernel\_size=(3, 3), strides=2, padding="same") \\
  BatchNormalization(momentum=0.8)                                         \\
  Activation("LeakyReLU", alpha=0.2)                                       \\
  Dropout(0.25)                                                            \\
  Flatten()                                                                \\
  Dense(1)                                                                 \\
  Activation("sigmoid")                                                    \\
  \hline
  \end{tabular}
\end{table}

\subsection{Learning phase}
\label{sec:learning-phase}

In this work, three methods are proposed for the learning phase: generator-only learning, GAN-only learning, and hybrid learning.
Generator-only learning learns the correlation between com\-press\-ed CSIs and images.
GAN-only learning uses both a generator and a discriminator. 
Hybrid learning combines the generator-only and GAN-only learning.

\subsection*{Generator-only learning}

\begin{figure}[b]
 \center
 \includegraphics[scale=0.25]{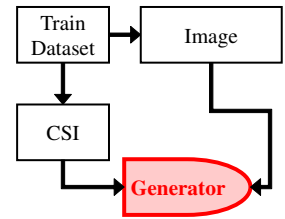}
 \caption{Generator-only learning}
\label{fig:g_only_learning}
\end{figure}

Figure~\ref{fig:g_only_learning} and Algorithm~\ref{alg:g_only_learning} depict the model structure and pseudo-code, respectively, of the generator-only learning. 
The con\-vo\-lu\-tional-neural-network-based generator is trained with the measured CSIs and the simultaneously captured images.
As generator-only learning learns the relations between the CSIs and images given as the training data, the generator may not accurately generate images from unknown CSIs.

\begin{algorithm}[b]
  \caption{Generator-only learning}
  \label{alg:g_only_learning}
  \begin{algorithmic}[1]
    \STATE $N$ $\Leftarrow$ number of training iterations
    \FOR{$i$ = 1 to $N$}
      \STATE csi\_list $\Leftarrow$ batch of CSI
      \STATE real\_images $\Leftarrow$ batch of real images
      \STATE $G$.csi2image.train(csi\_list, real\_images)
    \ENDFOR
  \end{algorithmic}
\end{algorithm}

\subsection*{GAN-only learning}

\begin{figure}[h]
 \center
 \includegraphics[scale=0.25]{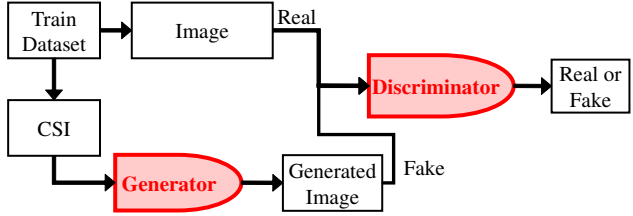}
 \caption{GAN-only learning}
\label{fig:dcgan_based_learning}
\end{figure}

Figure~\ref{fig:dcgan_based_learning} and Algorithm~\ref{alg:dcgan_based_learning} show the model structure and pseudo-code, respectively, of GAN-only learning.
Because the discriminator learns the converted image while judging whether or not it is a real image, this method is more likely to reconstruct a clear image than generator-only learning.
However, the discriminator only judges the legitimacy of the converted image, and it may not convert an image corresponding to the measured and com\-press\-ed CSI. 
In particular, the discriminator may not learn the detailed parts of the image.

\begin{algorithm}
  \caption{GAN-only learning}
  \label{alg:dcgan_based_learning}
  \begin{algorithmic}[1]
    \STATE $N$ $\Leftarrow$ number of training iterations
    \FOR{$i$ = 1 to $N$}
      \STATE csi\_list $\Leftarrow$ batch of CSI
      \STATE real\_images $\Leftarrow$ batch of real images
      \STATE $D$.train(real\_images, REAL)
      \STATE $D$.train($G$.generate(csi\_list), FAKE)
      \STATE $G$.generality.train(csi\_list, REAL)
    \ENDFOR
  \end{algorithmic}
\end{algorithm}

\subsection*{Hybrid learning}

\begin{figure}[bt]
 \center
 \includegraphics[scale=0.25]{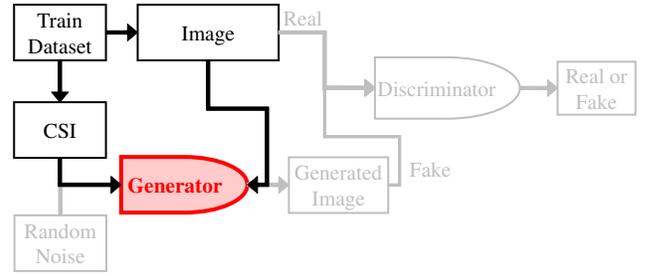}
 \caption{Hybrid learning: CSI2Image learning}
 \label{fig:csi2image_learning}
\end{figure}

\begin{figure}[bt]
 \center
 \includegraphics[scale=0.25]{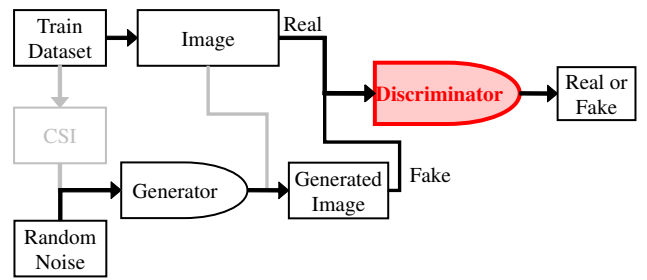}
 \caption{Hybrid learning: discriminator learning}
\label{fig:discriminator_learning}
\end{figure}

\begin{figure}[bt]
 \center
 \includegraphics[scale=0.25]{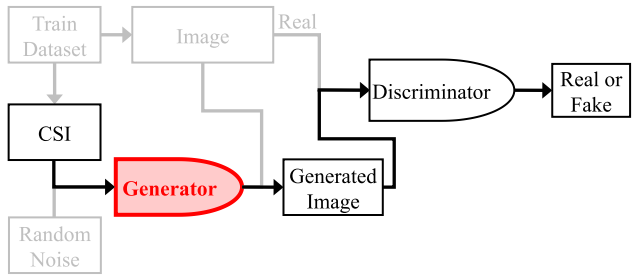}
 \caption{Hybrid learning: generality learning}
 \label{fig:generality_learning}
\end{figure}

Hybrid learning is the integration of generator-only learning and GAN-only learning.
The following four steps are regarded as one training epoch:
\begin{enumerate}
 \item CSI2Image learning (Figure \ref{fig:csi2image_learning})
 \item Image reconstruction via the generator
 \item Discriminator learning (Figure ~\ref{fig:discriminator_learning})
 \item Generality learning (Figure \ref{fig:generality_learning})
\end{enumerate}

Algorithm \ref{alg1} describes the pseudo-code of hybrid learning.
Let $N$ denote the number of training iterations and $K$ denote the interval for performing generality learning.
Lines 4 to 6 of Algorithm~\ref{alg1} represent CSI2Image learning.
CSI2Image learning obtains the training data and then trains the generator using the measured and compressed CSIs and the images corresponding to the CSIs at line 6, as shown in Figure~\ref{fig:csi2image_learning}.
Lines 7 to 8 of Algorithm~\ref{alg1} represent discriminator learning.
At line 7, the discriminator is trained to assess the training image to be real, and at line 8, it is trained to assess the generated image (obtained from the random noise) to be fake, as shown in Figure \ref{fig:discriminator_learning}.
Lines 9 to 11 of Algorithm~\ref{alg1} represent generality learning.
The generator is trained every $K$ epochs by feeding the compressed CSI to judge the generated image to be real by the discriminator, as shown in Figure \ref{fig:generality_learning}.  
When the value of $K$ is large, the generalization performance increases, while the CSI information is lost; when the value of $K$ is small, the image quality reduces because of generality loss.

\begin{algorithm}
  \caption{Hybrid Learning}
  \label{alg1}
  \begin{algorithmic}[1]
    \STATE $N$ $\Leftarrow$ number of training iterations
    \STATE $K$ $\Leftarrow$ interval of generality training
    \FOR{$i$ = 1 to $N$}
      \STATE csi\_list $\Leftarrow$ batch of CSI
      \STATE real\_images $\Leftarrow$ batch of real images
      \STATE $G$.csi2image.train(csi\_list, real\_images)
      \STATE $D$.train(real\_images, REAL)
      \STATE $D$.train($G$.generate(random\_noise), FAKE)
      \IF{$i \bmod K == 0$}
        \STATE $G$.generality.train(csi\_list, REAL)
      \ENDIF
    \ENDFOR
  \end{algorithmic}
\end{algorithm}

\subsection{Image generation phase}
\label{sec:generating-phase}
\begin{figure}[h]
 \center
 \includegraphics[scale=0.25]{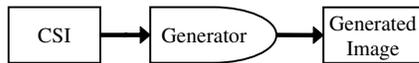}
 \caption{Image generation phase}
 \label{fig:mmgan_generating}
\end{figure}

In the image generation phase, the compressed CSI measured by wireless devices is fed into the pre-trained generator, and the generator converts the CSIs into full $64 \times 64$ pixel images, as shown in Figure \ref{fig:mmgan_generating}.

\section{Evaluation}
\label{sec:evaluation}

To clarify the effectiveness of the proposed CSI2Image, qualitative and quantitative evaluations were conducted.
Because the conversion of CSI to images is a new research area, no quantitative evaluation method has been established yet.
Therefore, a quantitative evaluation method using object detection and positional detection is proposed for the conversion of CSI to images.
The image converted from the CSI is applied to object detection, and the possibility of extracting the same detection results as the training image is evaluated.

\subsection{Evaluation settings}

\begin{figure}[bt]
  \centering
   \includegraphics[scale=0.3]{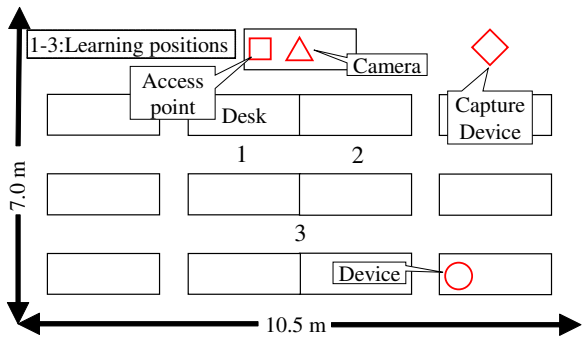}
   \caption{Experimental environment}
   \label{fig:a608}
\end{figure}

\begin{figure}[bt]
  \centering
   \includegraphics[scale=0.05]{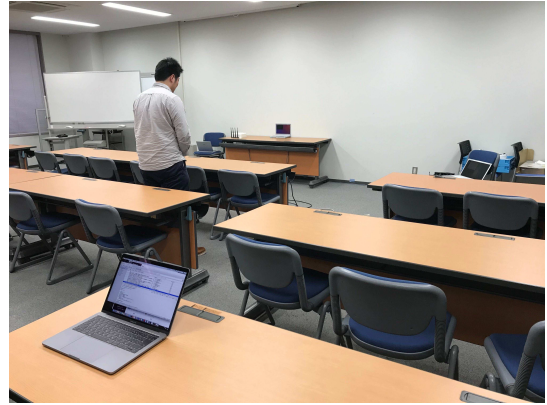}
   \caption{Snapshot of the experimental environment}
   \label{fig:a608real}
\end{figure}

\begin{figure*}[t]
  \begin{center}
   \centering
   \subfloat[Ground truth]{\label{fig:exp01_ok_base}\includegraphics[width=0.2\textwidth]{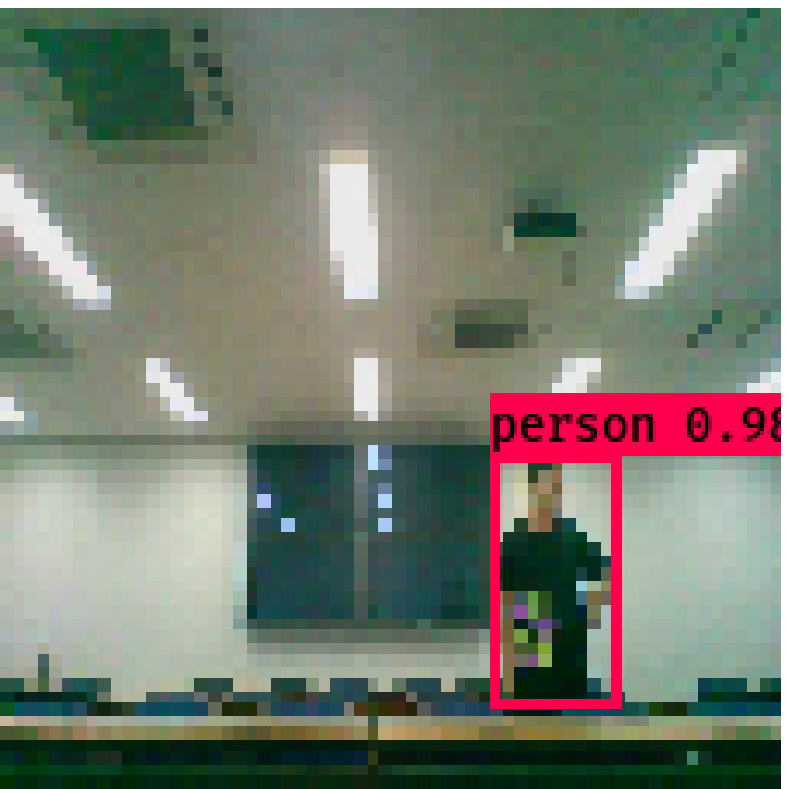}}
   \centering
   \hfill
   \subfloat[Generator-only learning]{\label{fig:exp01_ok_gonly}\includegraphics[width=0.2\textwidth]{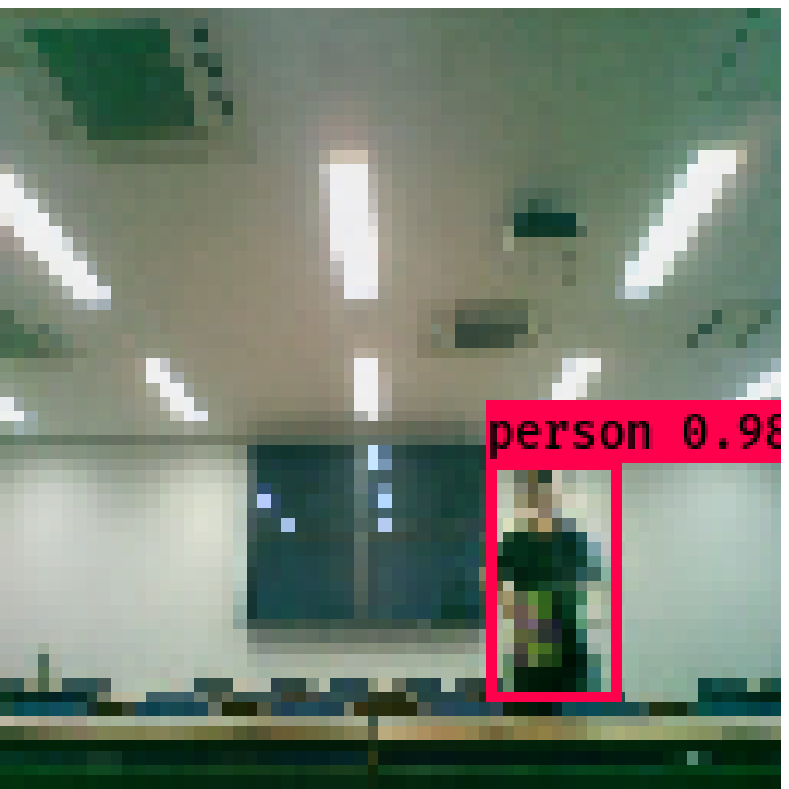}}
   \centering
   \hfill
   \subfloat[GAN-only learning]{\label{fig:exp01_ok_dcgan}\includegraphics[width=0.2\textwidth]{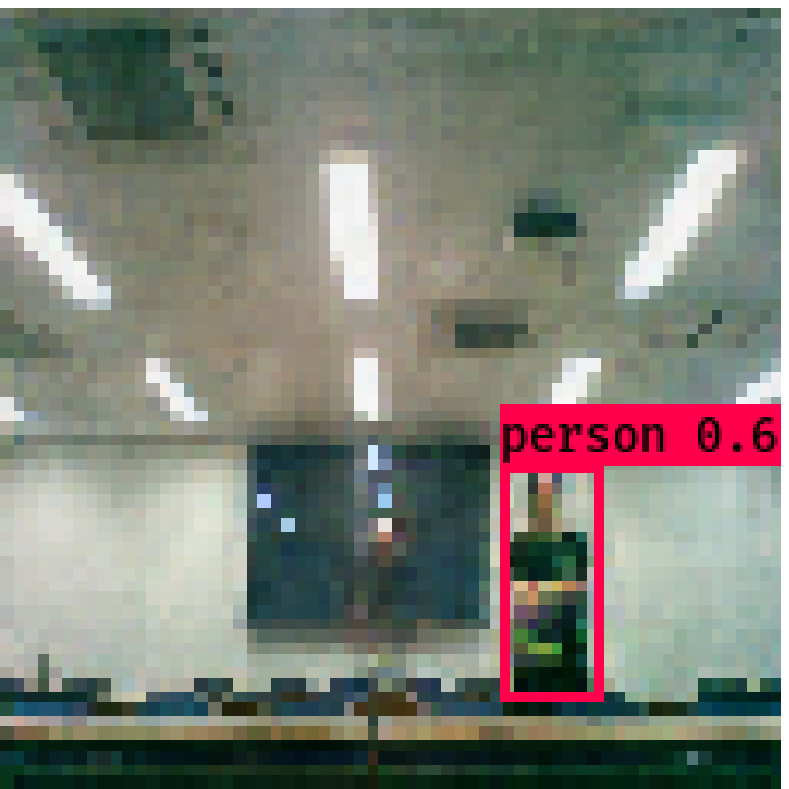}}
   \centering
   \hfill
   \subfloat[Hybrid learning]{\label{fig:exp01_ok_mmgan8}\includegraphics[width=0.2\textwidth]{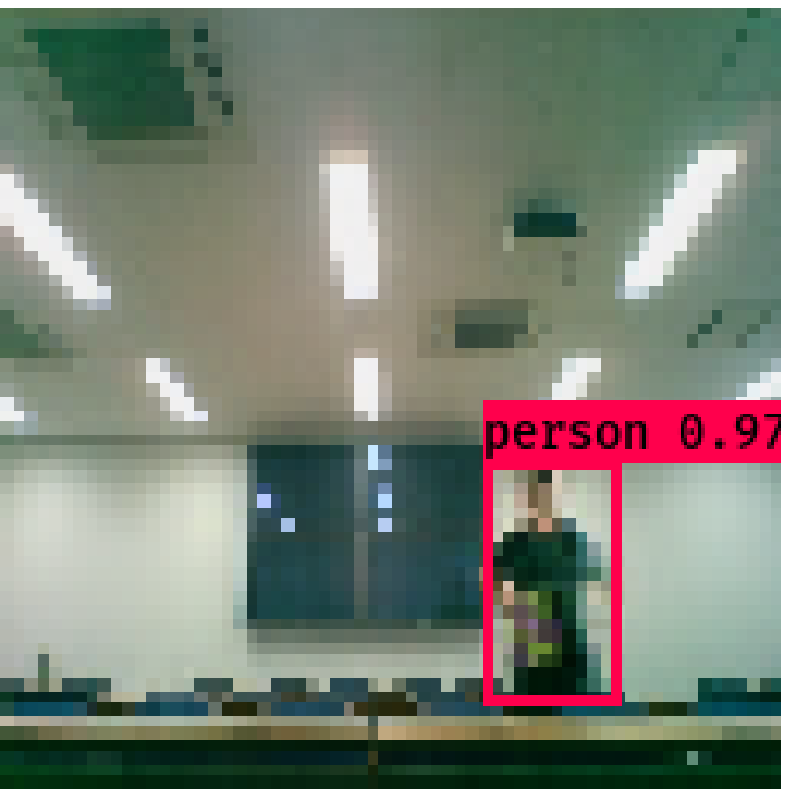}}
  \caption{Experiment 1: Examples of successful position detection with one user}
  \label{fig:exp01_ok}
  \end{center}
\end{figure*}

\begin{figure*}[t]
  \begin{center}
   \centering
   \subfloat[Ground truth]{\label{fig:exp01_ng_base}\includegraphics[width=0.2\textwidth]{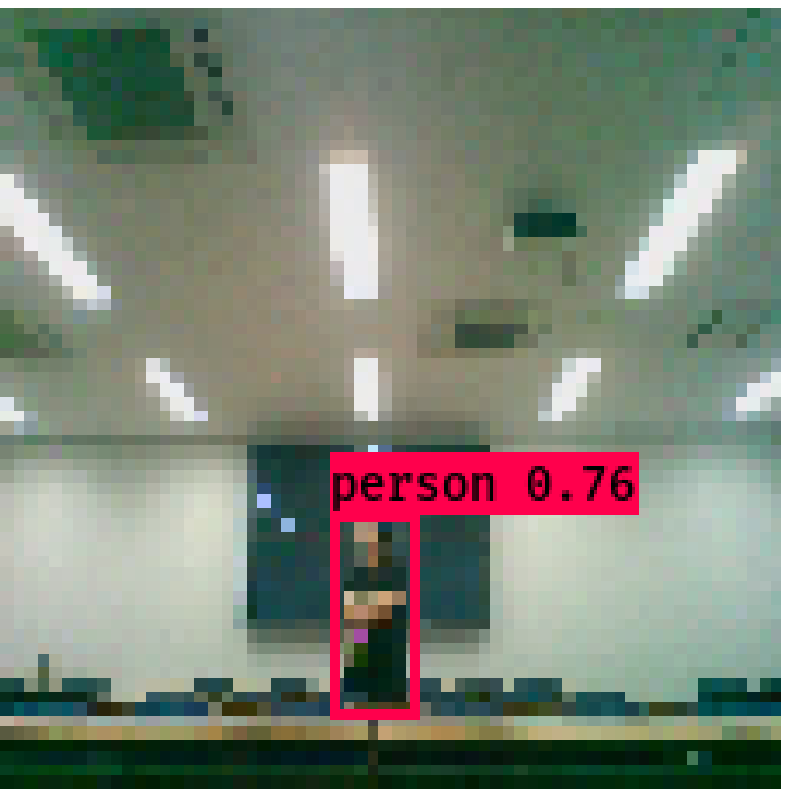}}
   \centering
   \hfill
   \subfloat[Generator-only learning]{\label{fig:exp01_ng_gonly}\includegraphics[width=0.2\textwidth]{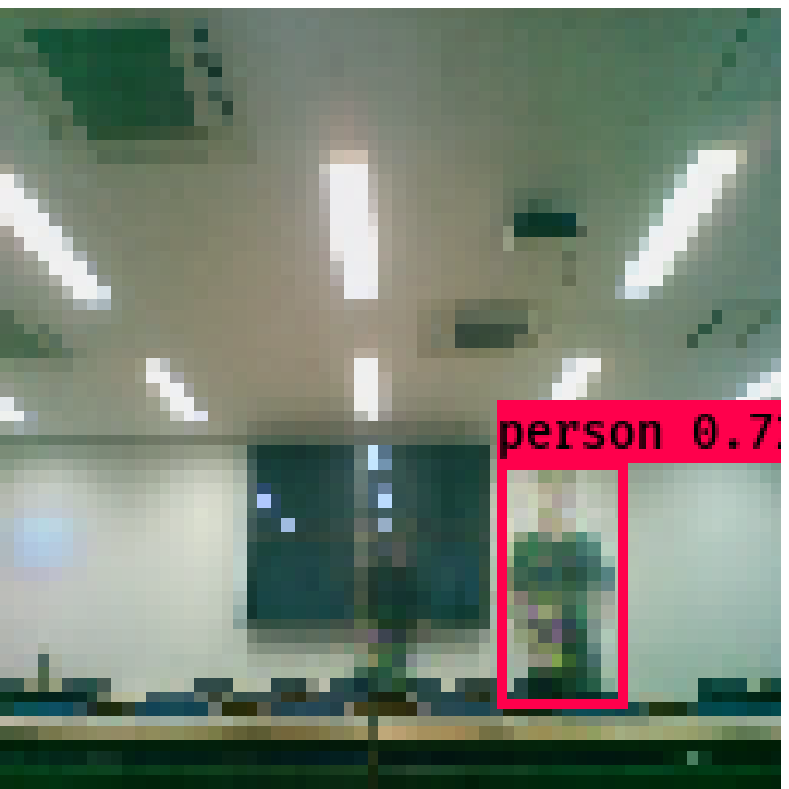}}
   \centering
   \hfill
   \subfloat[GAN-only learning]{\label{fig:exp01_ng_dcgan}\includegraphics[width=0.2\textwidth]{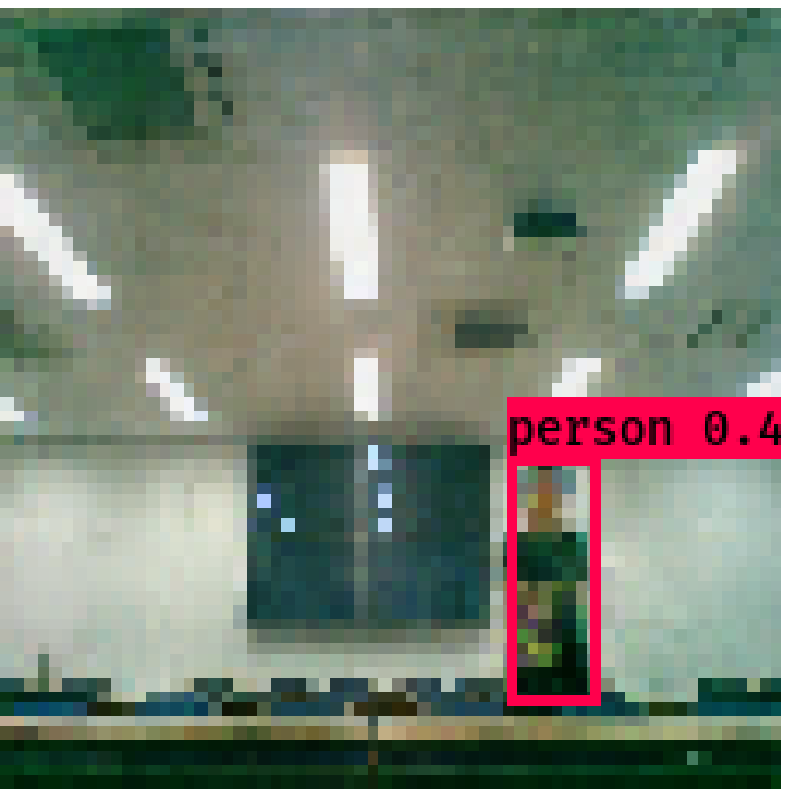}}
   \centering
   \hfill
   \subfloat[Hybrid learning]{\label{fig:exp01_ng_mmgan8}\includegraphics[width=0.2\textwidth]{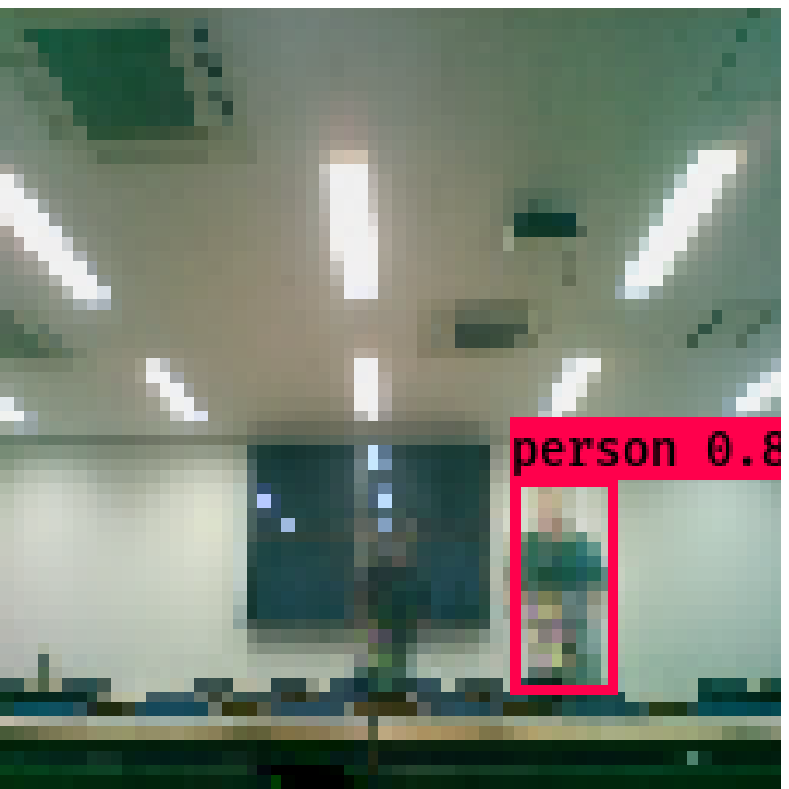}}
  \caption{Experiment 1: Examples of failed position detection with one user}
  \label{fig:exp01_ng}
  \end{center}
\end{figure*}

Figure~\ref{fig:a608} shows the configuration of each piece of equipment in the experimental environment, and Figure~\ref{fig:a608real} demonstrates a snapshot of the environment.
This experiment utilized an AP, a camera, a computer, and a capture device as a compressed CSI sniffer.
The AP was a Panasonic EA-7HW04AP1ES, the camera was a Panasonic CF-SX1GEPDR with a resolution of $1280 \times 720$ pixels, the computer was a MacBook Pro (13-inch, 2017), and the capture device was a Panasonic CF-B11\-QW\-H\-BR with CentOS 7.7-1908.
The proposed CSI2Image model was developed using a Dell Alienware 13 R3 computer, equipped with an Intel Core i7-7700HQ central processing unit, 16GB of DDR-SDRAM, a Geforce GTX 1060 graphics processing unit, and a solid state drive for storage.
The object detection library used was you only look once (YOLO) v3~\cite{redmon16:cvpr,yolov3}, and the model data were trained with \cite{yolov3.weight} from the COCO dataset~\cite{coco}.
The threshold to determine the object detection in YOLO is 0.3.

Three methods were compared: 
\begin{itemize}
 \item generator-only learning (gonly),
 \item GAN-only learning (gan), and
 \item hybrid learning (hybrid),
\end{itemize}
which are described in Section \ref{sec:learning-phase}.

In the quantitative evaluation, the following four aspects were extracted:
\begin{enumerate}
\item Object detection success rate. The high score obtained indicates that the quality of the generated images is sufficient for object detection. 
\item Average confidence score when object detection is successful. The confidence score is the confidence level of the object recognition algorithm in outputting the recognition result.
\item Structural similarity (SSIM) ~\cite{wang04:ieee}. SSIM is a standard measure of image quality which is used to evaluate the performance of image processing algorithms, such as image compression.
\item Position detection accuracy rate. The position detection accuracy rate is the percentage of correct locations detected via object recognition.
\end{enumerate}

The SSIM index is expressed by Equation ~\ref{equ:ssim},
\begin{equation}
  \label{equ:ssim}
  {\rm SSIM} = \frac{(2 \mu_x \mu_y + C_1)(2 \sigma_{xy} + C_2)}{(\mu_x^2 + \mu_y^2 + C_1)(\sigma_x^2 + \sigma_y^2 + C_2)}
\end{equation}
where $x$ and $y$ are vectors, whose elements are the pixels of an original image and the reconstructed images, respectively.
Let $\mu_x$ and $\mu_y$ denote the average pixel values of images $x$ and $y$, $\sigma_x$ and $\sigma_y$ be standard deviations of images $x$ and $y$, and $\sigma_{xy}$ be the covariance of images $x$ and $y$.
Both $C_1$ and $C_2$ are constant values defined as $C_1 = (255K_1)^2$ and $C_2 = (255K_2)^2$, respectively.
In this case, the parameters of $K_1 = 0.01$ and $K_2 = 0.03$ are the same values as in \cite{wang04:ieee}.
The SSIM index takes a value from 0 to 1, where 1 represents an exact image match.

\subsection{Experiment 1: Single-user position detection}
\label{sec:eval-user-1}

To clarify the baseline performance of the proposed CSI2Image, single-user location detection was evaluated.
The experiment was performed with only one person at positions 1 to 3 in Figure ~\ref{fig:a608}.
Three types of image patterns were possible, in which the person would be at position 1, 2, or 3, respectively.
The evaluation used 180 images as training data and 184 images as test data.
The number of epochs was 32,000, and the batch size was 32.
In hybrid learning, $K$ was eight.

\subsection*{Qualitative evaluation}

Figure \ref{fig:exp01_ok} shows an example of successful position detection with one user.
The red square on each figure represents the object detection results obtained using YOLO.
If a person is detected on the right of the image, as shown in Figure \ref{fig:exp01_ok}\subref{fig:exp01_ok_base}, the position detection is accurate.
The positions of generator-only learning and hybrid learning in Figures \ref{fig:exp01_ok}\subref{fig:exp01_ok_gonly} and \ref{fig:exp01_ok}\subref{fig:exp01_ok_mmgan8} are accurate, as is the shape of the person.
On the other hand, GAN-only learning, shown in Figure \ref{fig:exp01_ok}\subref{fig:exp01_ok_dcgan}, accurately detects the position of the person, while a shadow of the person is also output in the center of the incorrect position.

Figure \ref{fig:exp01_ng} shows an example of failed position detection with one user.
If a person is detected on the right of the image, as shown in Figure~\ref{fig:exp01_ng}\subref{fig:exp01_ng_base}, the position detection is accurate.
As can be seen from Figures \ref{fig:exp01_ng}\subref{fig:exp01_ng_gonly} to \ref{fig:exp01_ng}\subref{fig:exp01_ng_mmgan8}, pale ghost-like shadows appear at the middle and the right of the images.
In contrast, GAN-only learning in Figure~\ref{fig:exp01_ng}\subref{fig:exp01_ng_dcgan} produces a clean image as compared to generator-only learning and hybrid learning, although the position is inaccurate.

\subsection*{Quantitative evaluation}
\label{sec:exp1-quantitative}

\begin{figure*}[t]
  \begin{center}
   \centering
   \subfloat[Successful detection rate]{\label{fig:exp-01-01_drate_32000}\includegraphics[width=0.25\textwidth]{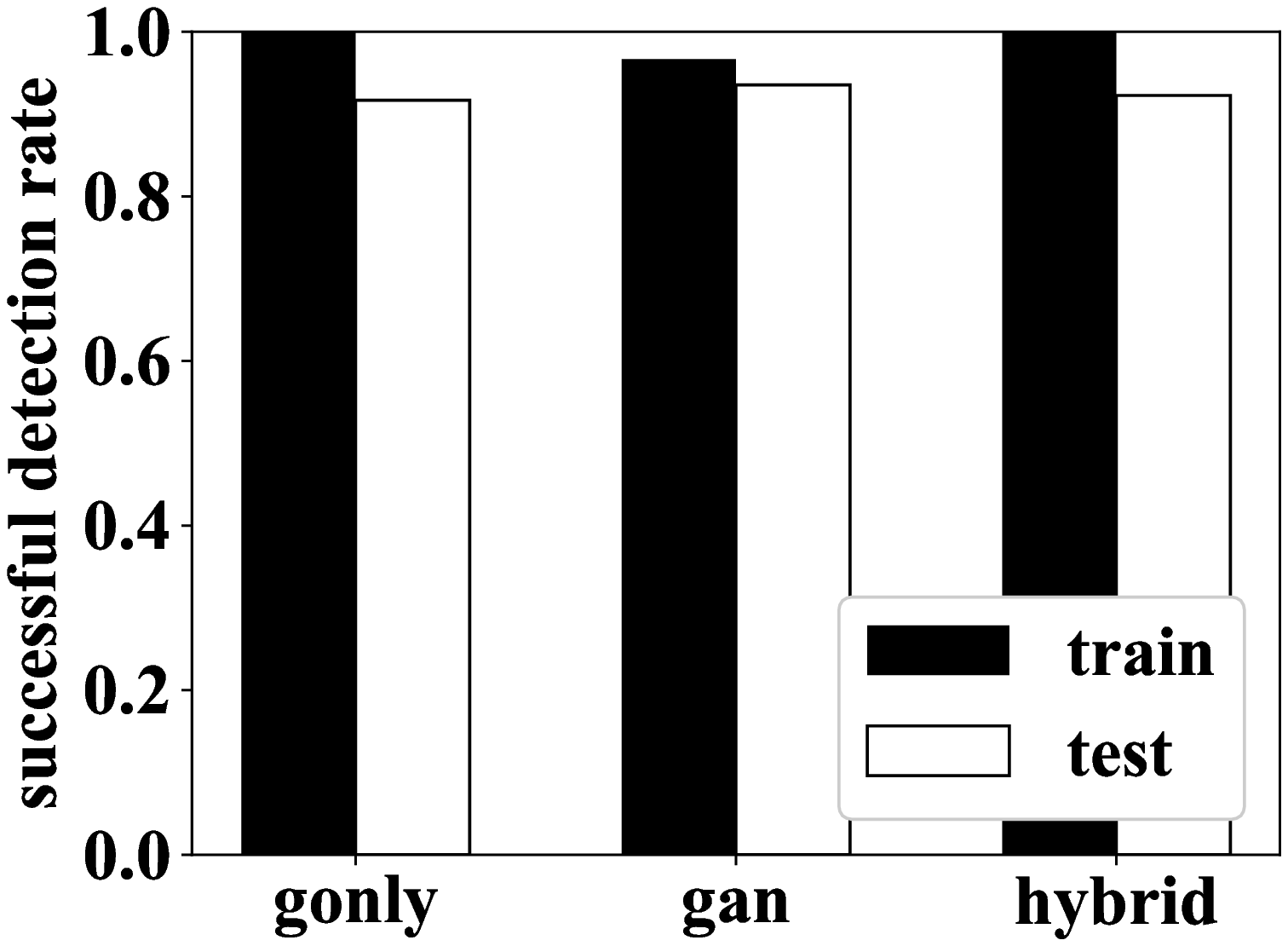}}
   \centering
   \hfill
   \subfloat[Average confidence score]{\label{fig:exp-01-01_cscore_32000}\includegraphics[width=0.25\textwidth]{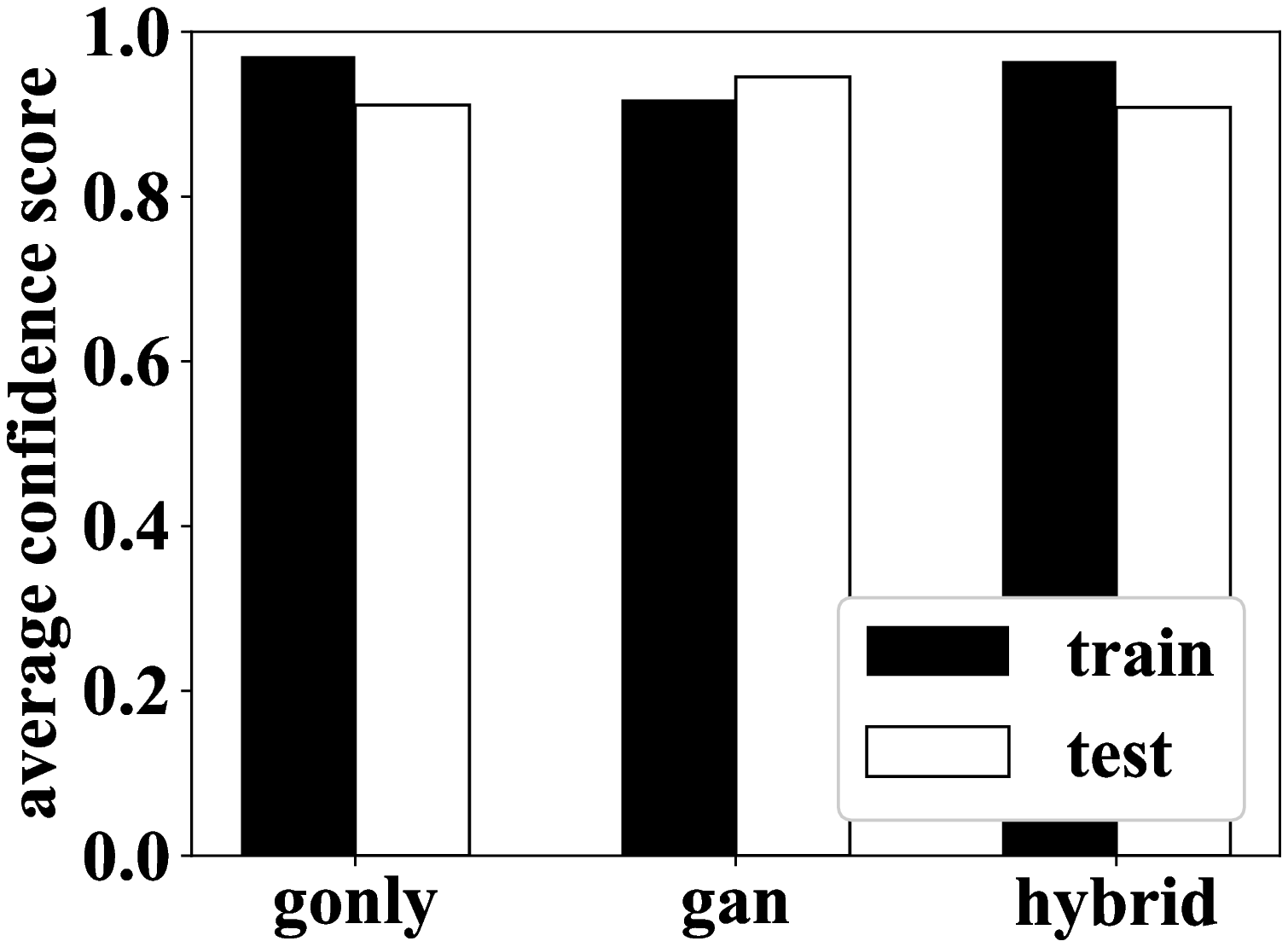}}
   \centering
   \hfill
   \subfloat[SSIM index]{\label{fig:exp-01-01_ssim_32000}\includegraphics[width=0.25\textwidth]{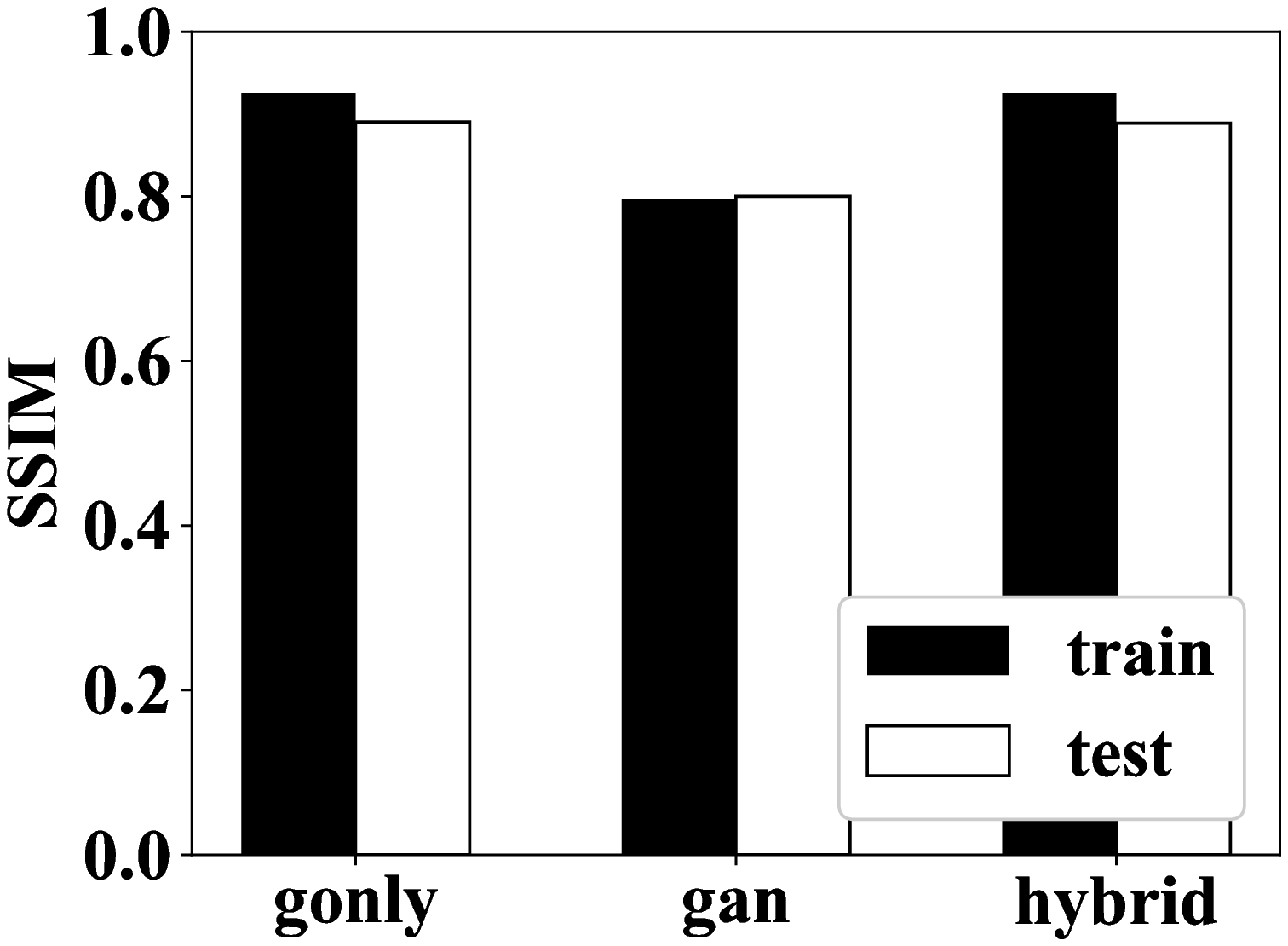}}
   \centering
   \hfill
   \subfloat[Accuracy]{\label{fig:exp-01-01_32000}\includegraphics[width=0.25\textwidth]{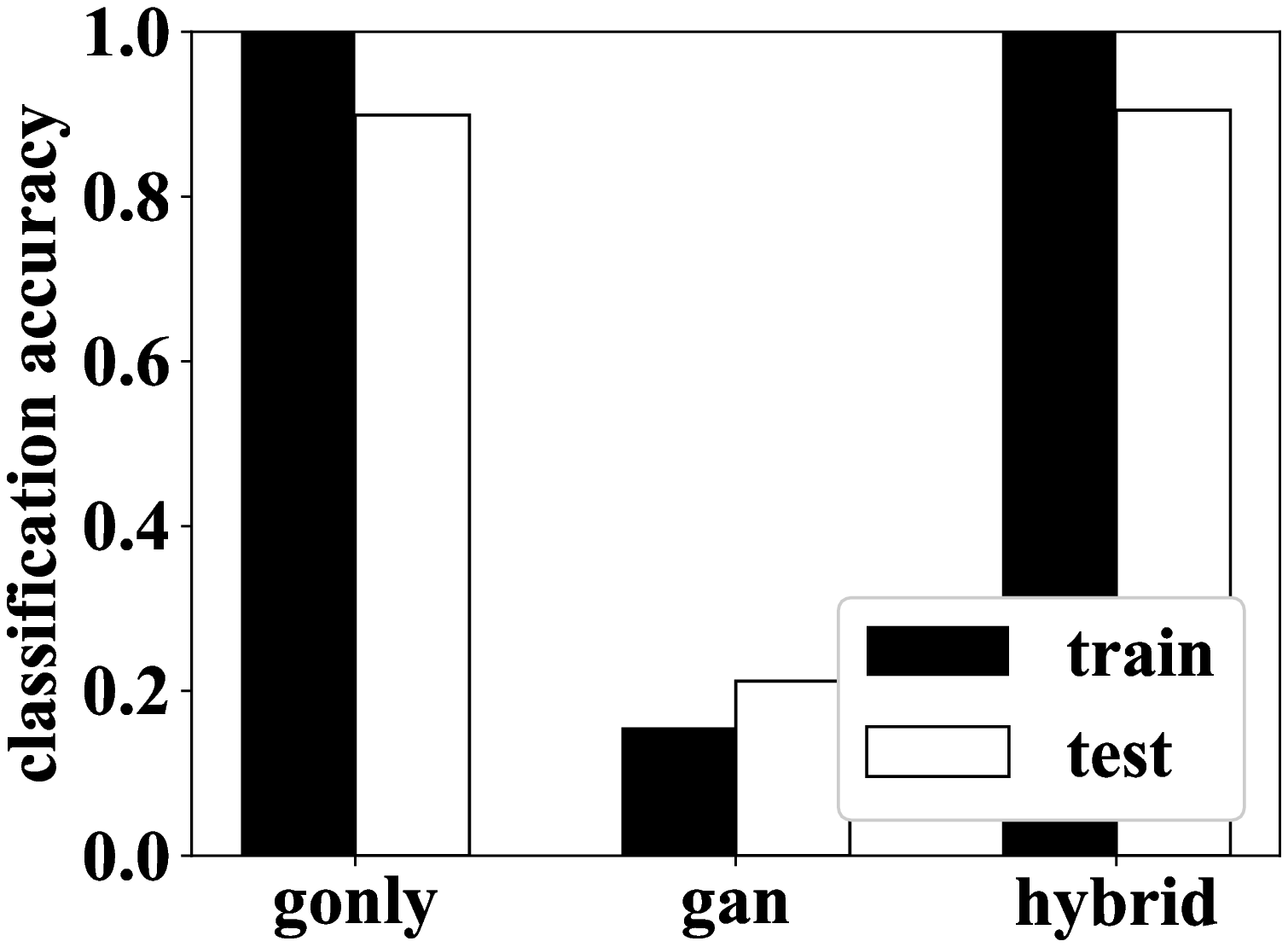}}
  \caption{Experiment 1: Quantitative evaluation of single-user position classification}
  \label{fig:exp01-quantitative}
  \end{center}
\end{figure*}

\begin{figure*}[t]
  \begin{center}
   \centering
   \subfloat[Ground truth]{\label{fig:exp03_ok_base}\includegraphics[width=0.2\textwidth]{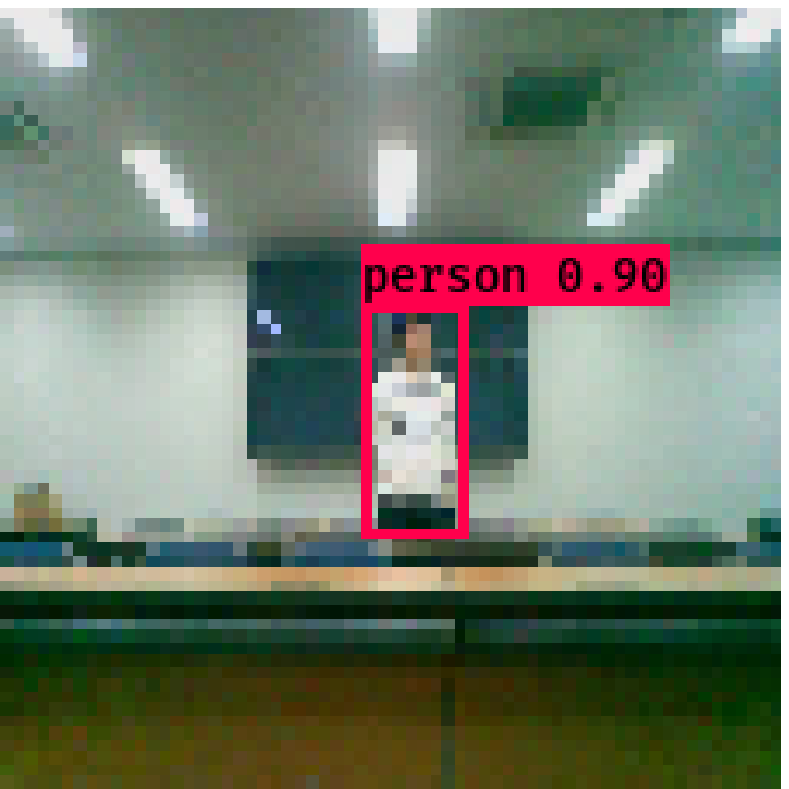}}
   \centering
   \hfill
   \subfloat[Generator-only learning]{\label{fig:exp03_ok_gonly}\includegraphics[width=0.2\textwidth]{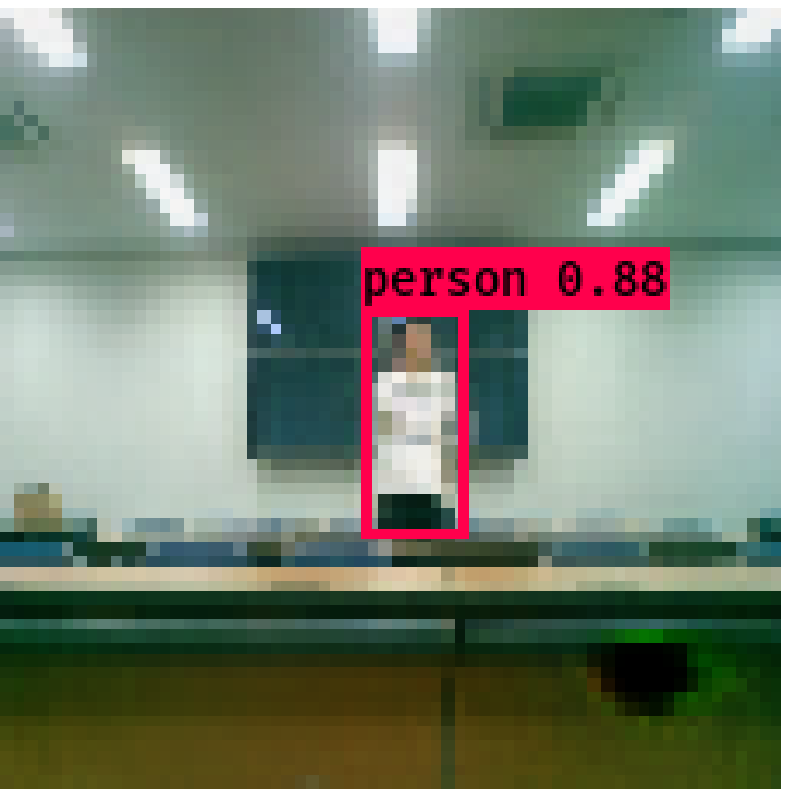}}
   \centering
   \hfill
   \subfloat[GAN-only learning]{\label{fig:exp03_ok_dcgan}\includegraphics[width=0.2\textwidth]{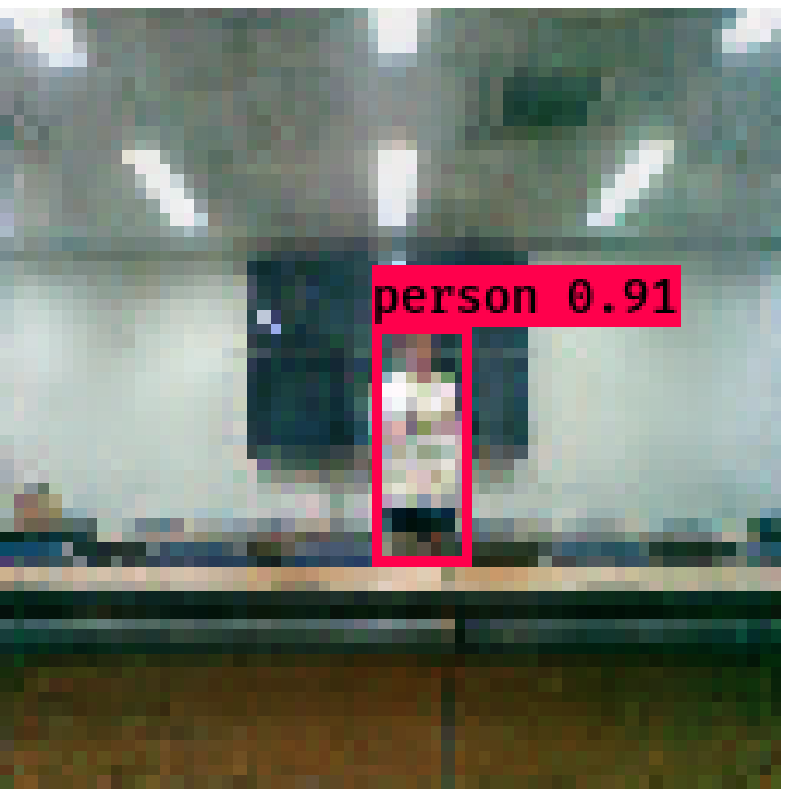}}
   \centering
   \hfill
   \subfloat[Hybrid learning]{\label{fig:exp03_ok_mmgan8}\includegraphics[width=0.2\textwidth]{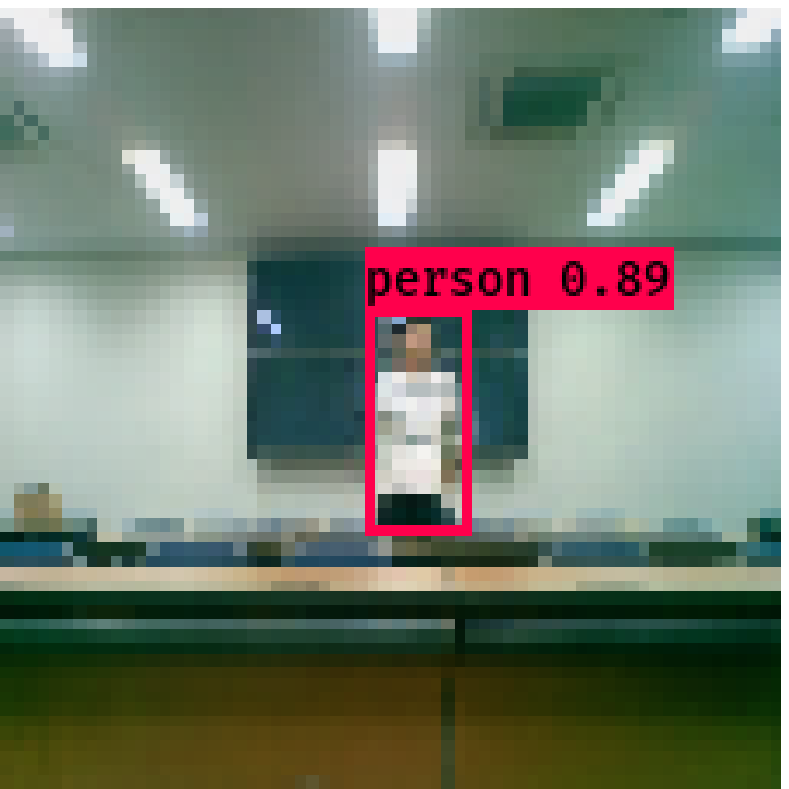}}
  \caption{Experiment 2: Examples of successful position detection with one or two users}
  \label{fig:exp03_ok}
  \end{center}
\end{figure*}

\begin{figure*}[t]
  \begin{center}
   \centering
   \subfloat[Ground truth]{\label{fig:exp03_ng_base}\includegraphics[width=0.2\textwidth]{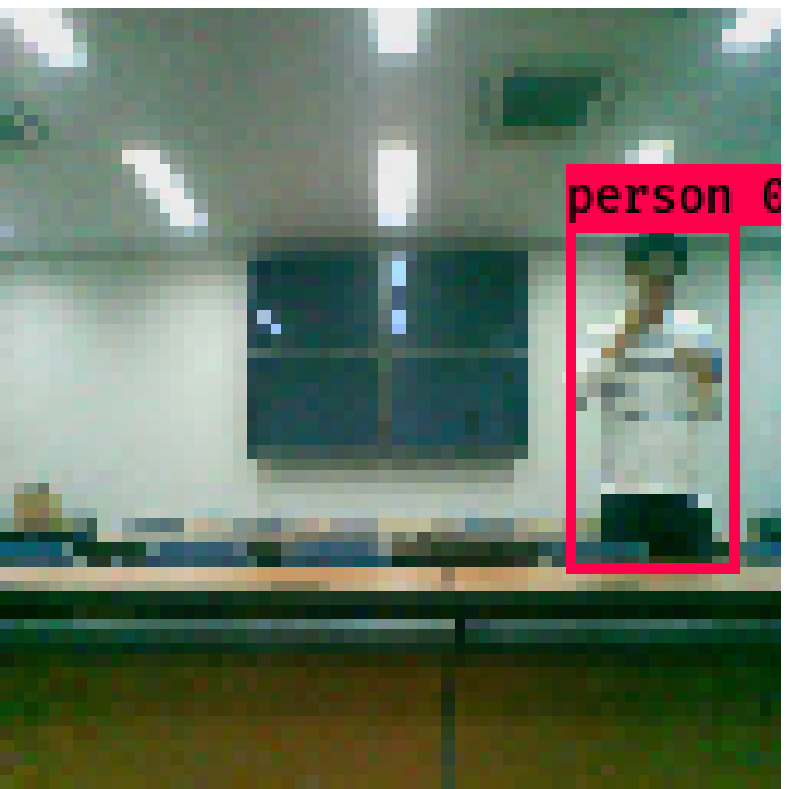}}
   \centering
   \hfill
   \subfloat[Generator-only learning]{\label{fig:exp03_ng_gonly}\includegraphics[width=0.2\textwidth]{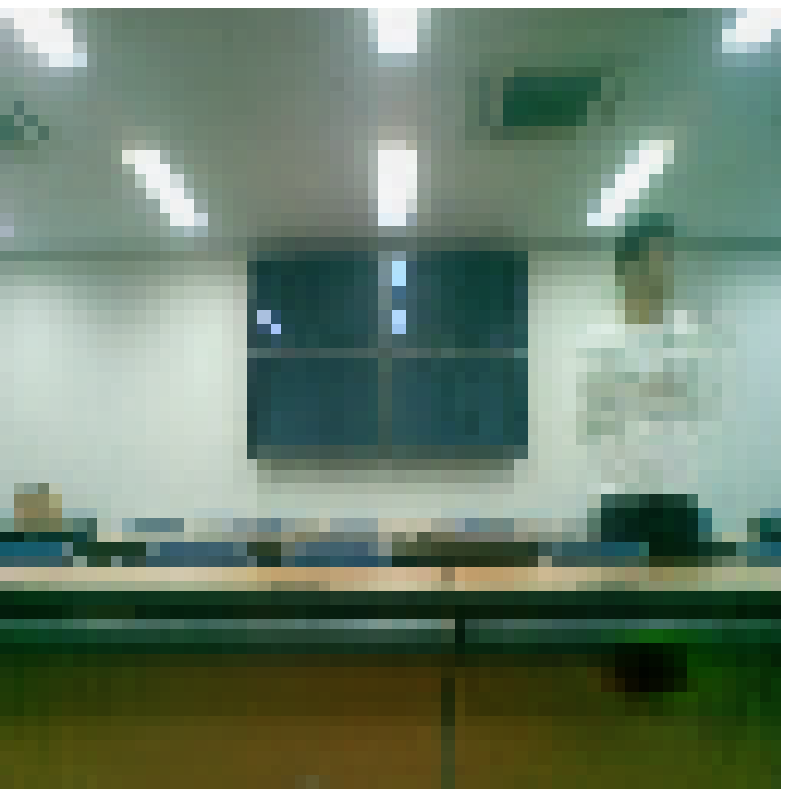}}
   \centering
   \hfill
   \subfloat[GAN-only learning]{\label{fig:exp03_ng_dcgan}\includegraphics[width=0.2\textwidth]{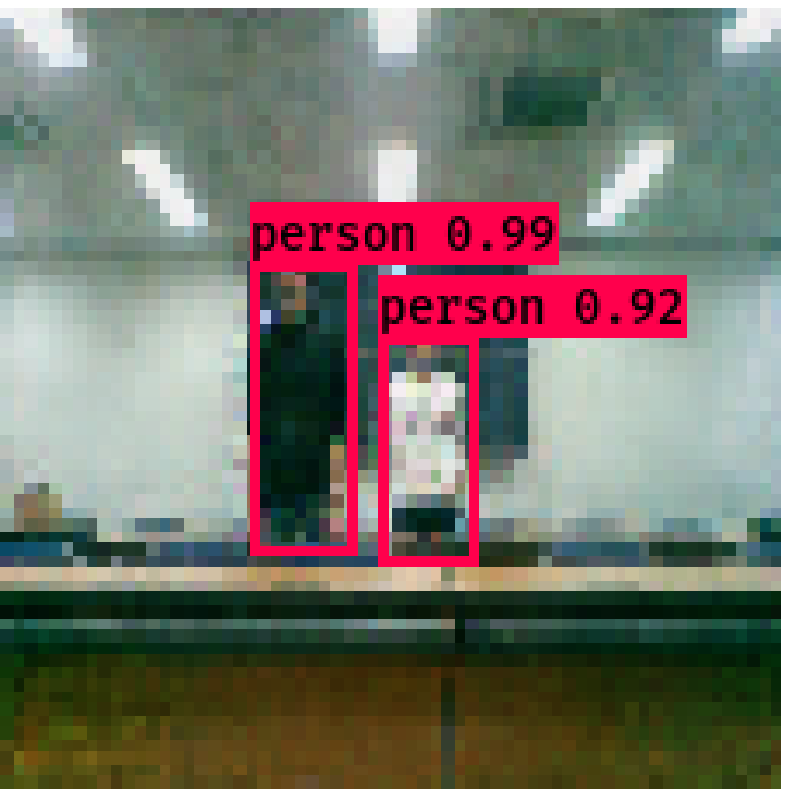}}
   \centering
   \hfill
   \subfloat[Hybrid learning]{\label{fig:exp03_ng_mmgan8}\includegraphics[width=0.2\textwidth]{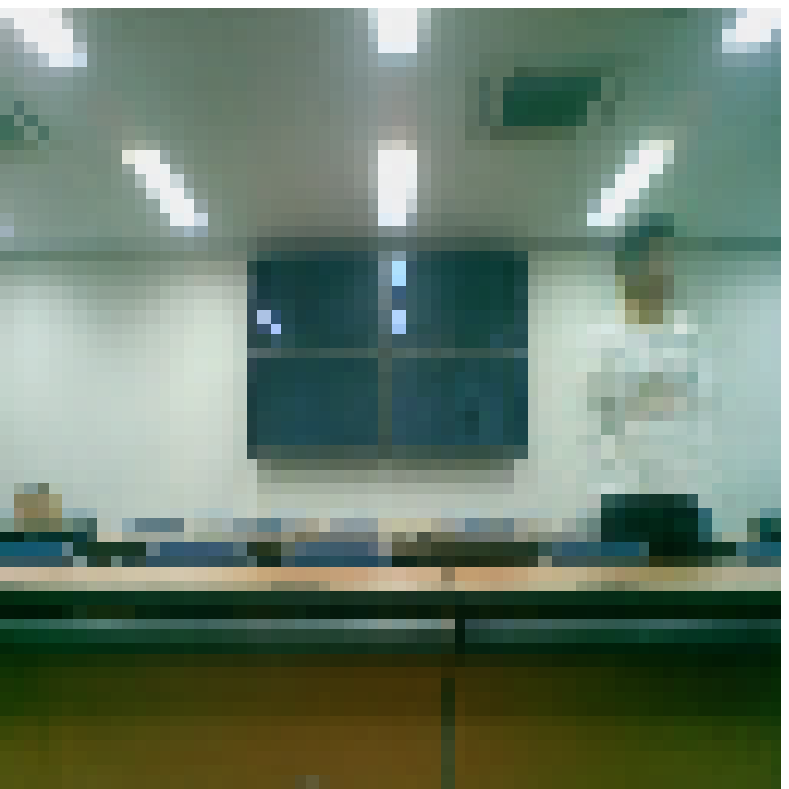}}
  \caption{Experiment 2: Examples of failed position detection with one or two users}
  \label{fig:exp03_ng}
  \end{center}
\end{figure*}

Figure~\ref{fig:exp01-quantitative}\subref{fig:exp-01-01_drate_32000} shows the success rate of human detection.
The black and white bars represent the results using the training and test data, respectively.
The confidence threshold of YOLO is 0.3.
In terms of the detection success rate, GAN-only learning achieved the highest score: in the test data, the detection success rate of generator-only learning, GAN-only learning, and hybrid learning were approximately 92.7 \%, 93.5 \%, and 92.3 \%, respectively.

Figure~\ref{fig:exp01-quantitative}\subref{fig:exp-01-01_cscore_32000} shows the average confidence score when the object detection is successful.
In addition to the detection successful rate described above, GAN-only learning achieved the highest score: in the test data, the average confidence score of generator-only learning, GAN-only learning, and hybrid learning were approximately 91.1 \%, 94.5 \%, and 90.8 \%, respectively.

Figure~\ref{fig:exp01-quantitative}\subref{fig:exp-01-01_ssim_32000} shows the SSIM index of each comparison method.
In contrast to the detection success rates and the average confidence score, GAN-only learning showed the worst performance in terms of the SSIM index.
With the test data, the results of the SSIM index were approximately 0.890 for generator-only learning, 0.800 for GAN-only learning, and 0.889 for hybrid learning.

To understand the reason for the low SSIM performance of GAN-only learning, the position detection accuracy was evaluated.
The results show that GAN-only learning had the worst performance compared to generator-only learning and hybrid learning.
With the test data, the accuracies were found to be approximately 89.9 \% for generator-only learning, 21.2 \% for GAN-only learning, and 90.5 \% for hybrid learning.

Thus, although the detection success rate and the average confidence score are the highest in GAN-only learning, the SSIM index is low owing to misplaced-user images.
In particular, GAN-only learning has a position detection accuracy even with the training data.
This is because GAN-only learning only learns the legitimacy of the generated image using the discriminator, as shown in Figure~\ref{fig:exp01-quantitative}\subref{fig:exp-01-01_32000}.

\begin{figure*}[t]
  \begin{center}
   \centering
   \subfloat[Detection success rate]{\label{fig:exp-03-01_drate_32000}\includegraphics[width=0.25\textwidth]{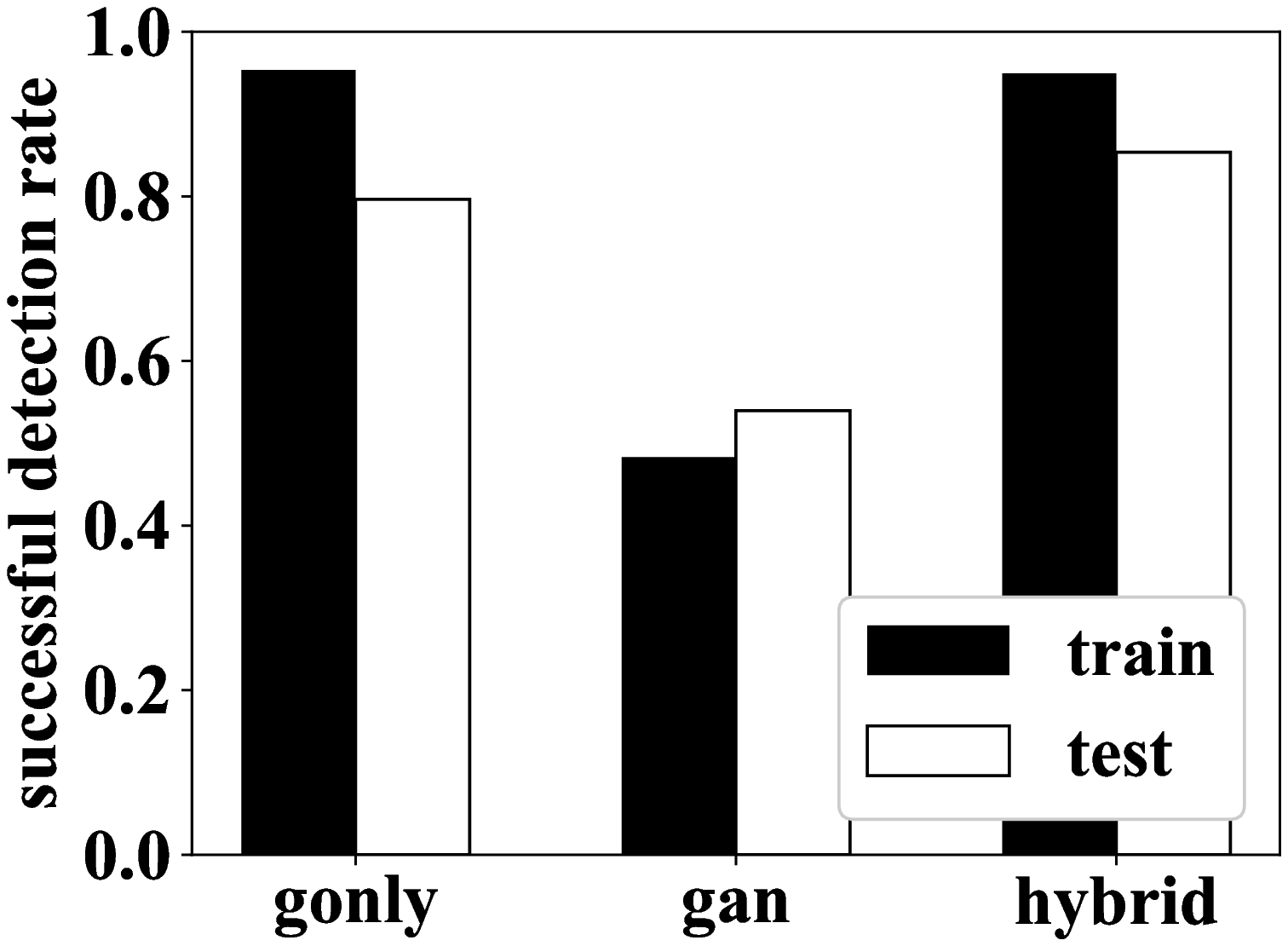}}
   \centering
   \hfill
   \subfloat[Average confidence score]{\label{fig:exp-03-01_cscore_32000}\includegraphics[width=0.25\textwidth]{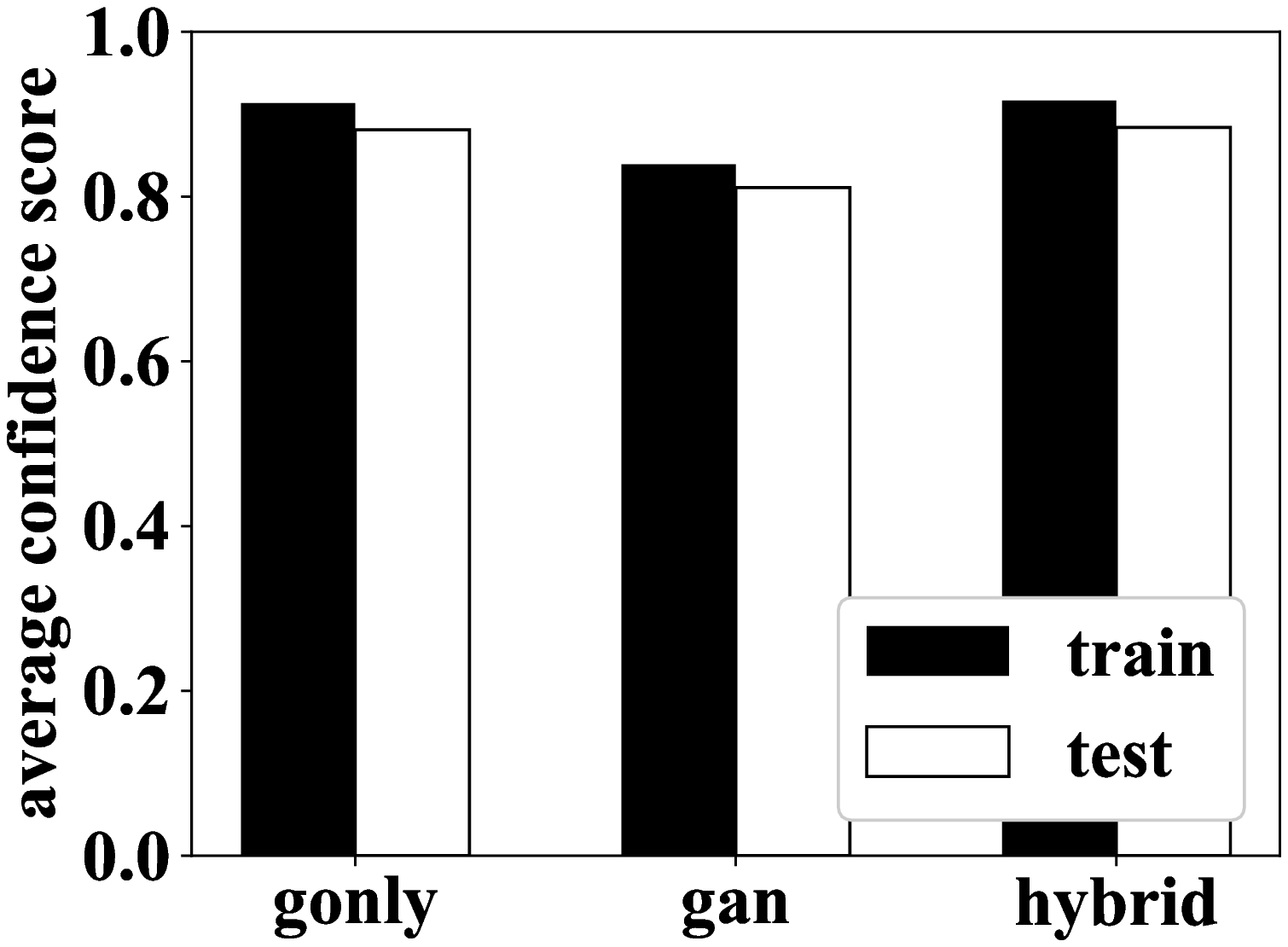}}
   \centering
   \hfill
   \subfloat[SSIM index]{\label{fig:exp-03-01_ssim_32000}\includegraphics[width=0.25\textwidth]{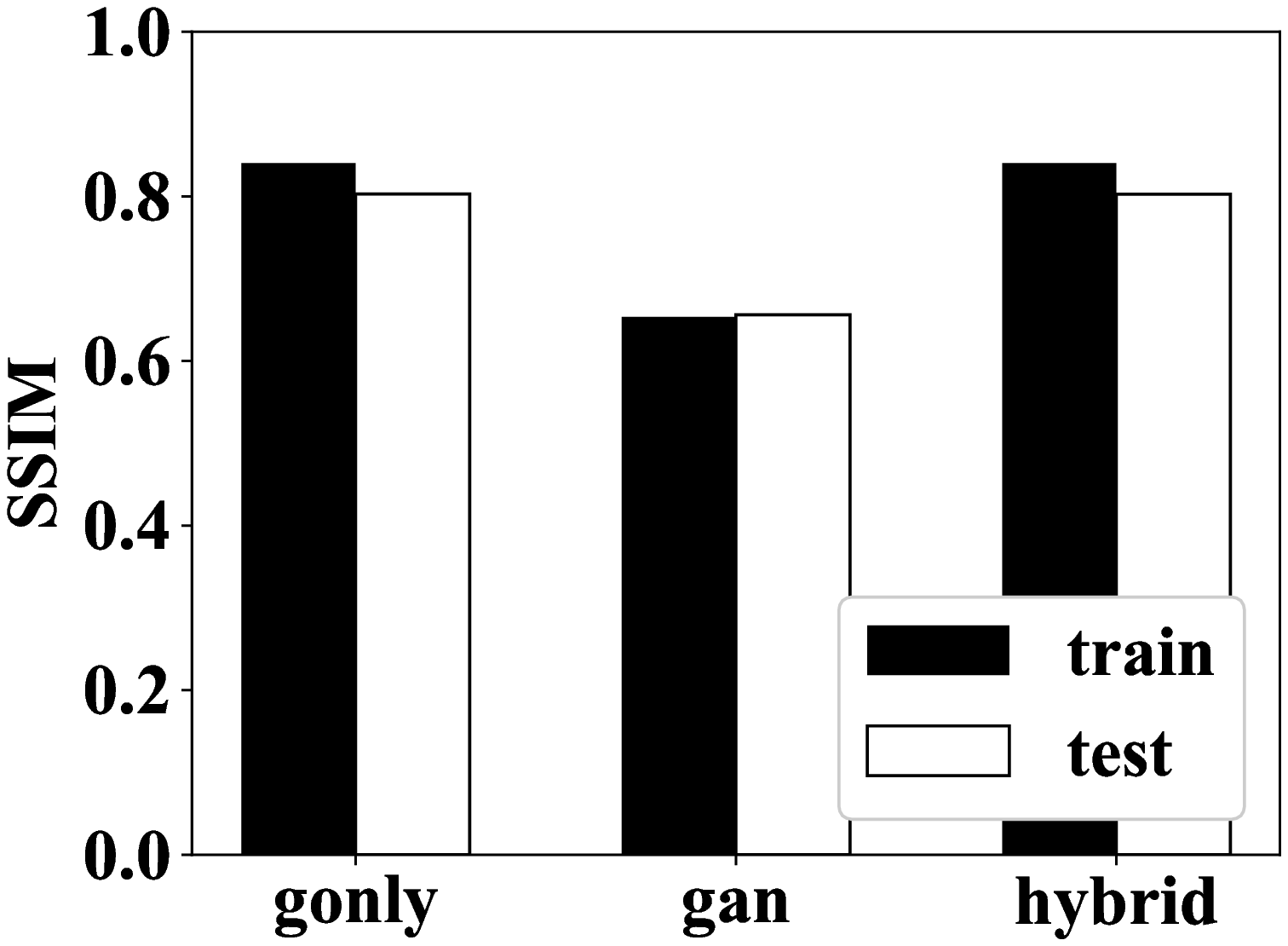}}
   \centering
   \hfill
   \subfloat[Accuracy]{\label{fig:exp-03-01_32000}\includegraphics[width=0.25\textwidth]{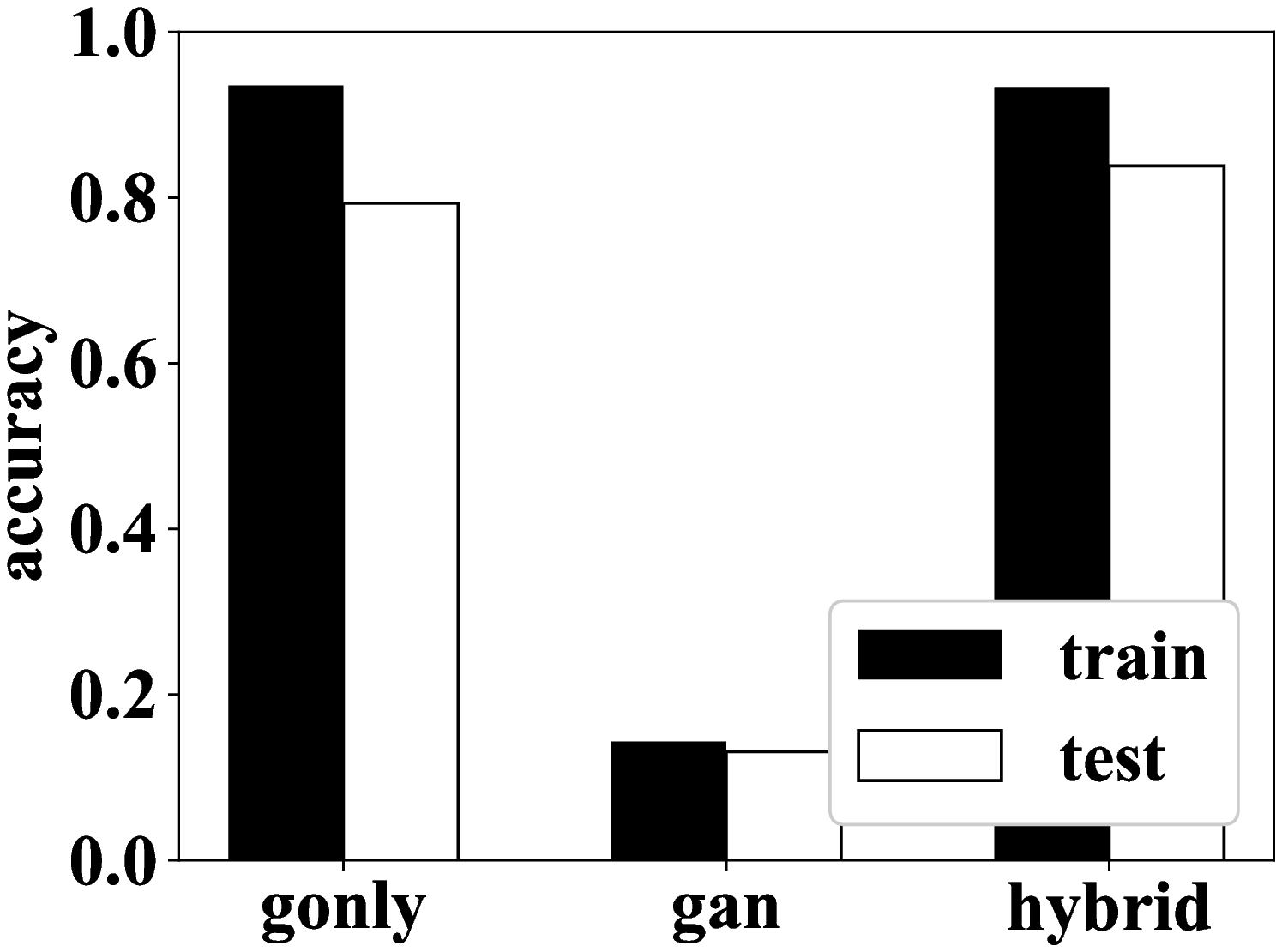}}
  \caption{Experiment 2: Quantitative evaluation of the position classification of one or two users}
  \label{fig:exp02-quantitative}
  \end{center}
\end{figure*}

\subsection{Experiment 2: Position detection for one or two users}
\label{sec:eval-user-2}

For more complex situations, the position detection was evaluated for the case of one or two users.
Specifically, six types of classification problems were evaluated when one person or two people were at positions 1 to 3, as shown in Figure~\ref{fig:a608}: ``one person at 1,'' ``one person at 2,'' ``one person at 3,'' ``two people at 1 and 2,'' ``two people at 1 and 3,'' and ``two people at 2 and 3.''
We used 720 images as training data and 330 images as test data for the evaluation.
The other conditions were identical to those presented in Section \ref{sec:eval-user-1}.

\subsection*{Qualitative evaluation}

Figure \ref{fig:exp03_ok} shows an example of successful position detection with one or two users.
If a person can be detected at the center of the image, as shown in Figure \ref{fig:exp03_ok}\subref{fig:exp03_ok_base}, the position detection is accurate.
The positions obtained via generator-only learning and hybrid learning in Figures \ref{fig:exp03_ok}\subref{fig:exp03_ok_gonly} and \ref{fig:exp03_ok}\subref{fig:exp03_ok_mmgan8} are accurate, and the human shape is clearly displayed.
However, as shown in in Figure \ref{fig:exp03_ok}\subref{fig:exp03_ok_dcgan}, GAN-only learning accurately detects the position, but a shad\-ow is also output on the left side of the incorrect position.

Figure \ref{fig:exp03_ng} shows an example of failed position detection with one or two users.
If a person is detected on the right side of the image, as shown in Figure~\ref{fig:exp03_ng}\subref{fig:exp03_ng_base}, the position detection is accurate.
As shown in Figures~\ref{fig:exp03_ng}\subref{fig:exp03_ng_gonly} and \ref{fig:exp03_ng}\subref{fig:exp03_ng_mmgan8}, YOLO does not detect the person on the right side of the image for generator-only learning and hybrid learning, although human-like objects are displayed on the right side.
In addition, GAN-only learning shows people in the middle and left of the incorrect position, as shown in Figure \ref{fig:exp03_ng}\subref{fig:exp03_ng_dcgan}.

\subsection*{Quantitative evaluation}
\label{sec:exp2-quantitative}

Figure~\ref{fig:exp02-quantitative}\subref{fig:exp-03-01_drate_32000} shows the successful detection rates of each comparison method.
It can be observed that hybrid learning achieves the highest detection rate: using the test data, the detection success rates are approximately 79.6 \% for generator-only learning, 54.0 \% for GAN-only learning, and 85.4 \% for hybrid learning.
The detection success rate of GAN-only learning is low even when using training data.

Figure ~\ref{fig:exp02-quantitative}\subref{fig:exp-03-01_cscore_32000} shows the average confidence score of each comparison method.
Similar to the above successful detection rate, it can be observed that the confidence score of hybrid learning is the highest: using the test data, the average confidence scores are approximately 88.1 \% for generator-only learning, 81.1 \% for GAN-only learning, and 88.4 \% for hybrid learning.

Figure~\ref{fig:exp02-quantitative}\subref{fig:exp-03-01_ssim_32000} shows the SSIM index of each comparison method.
The results are the same as in the single-user evaluation: GAN-only learning shows the worst performance.
Using the test data, the SSIM indexes are 0.803 for generator-only learning, 0.656 for GAN-only learning, and 0.803 for hybrid learning.

Figure ~\ref{fig:exp02-quantitative}\subref{fig:exp-03-01_32000} shows the position detection accuracy of each comparison method.
The results show that using the test data, hybrid learning achieved the highest accuracy, while GAN-only learning had the lowest: the values were  79.3\% for generator-only learning, 13.1\% for GAN-only learning, and 83.8\% for hybrid learning.

\subsection{Experiment 3: Continuous Position Estimation for a Single User}

\begin{figure}[tb]
  \begin{center}
   \hspace{-5mm}
   \subfloat[Detection success rate]{\label{fig:exp-02-01_drate_32000}\includegraphics[width=0.25\textwidth]{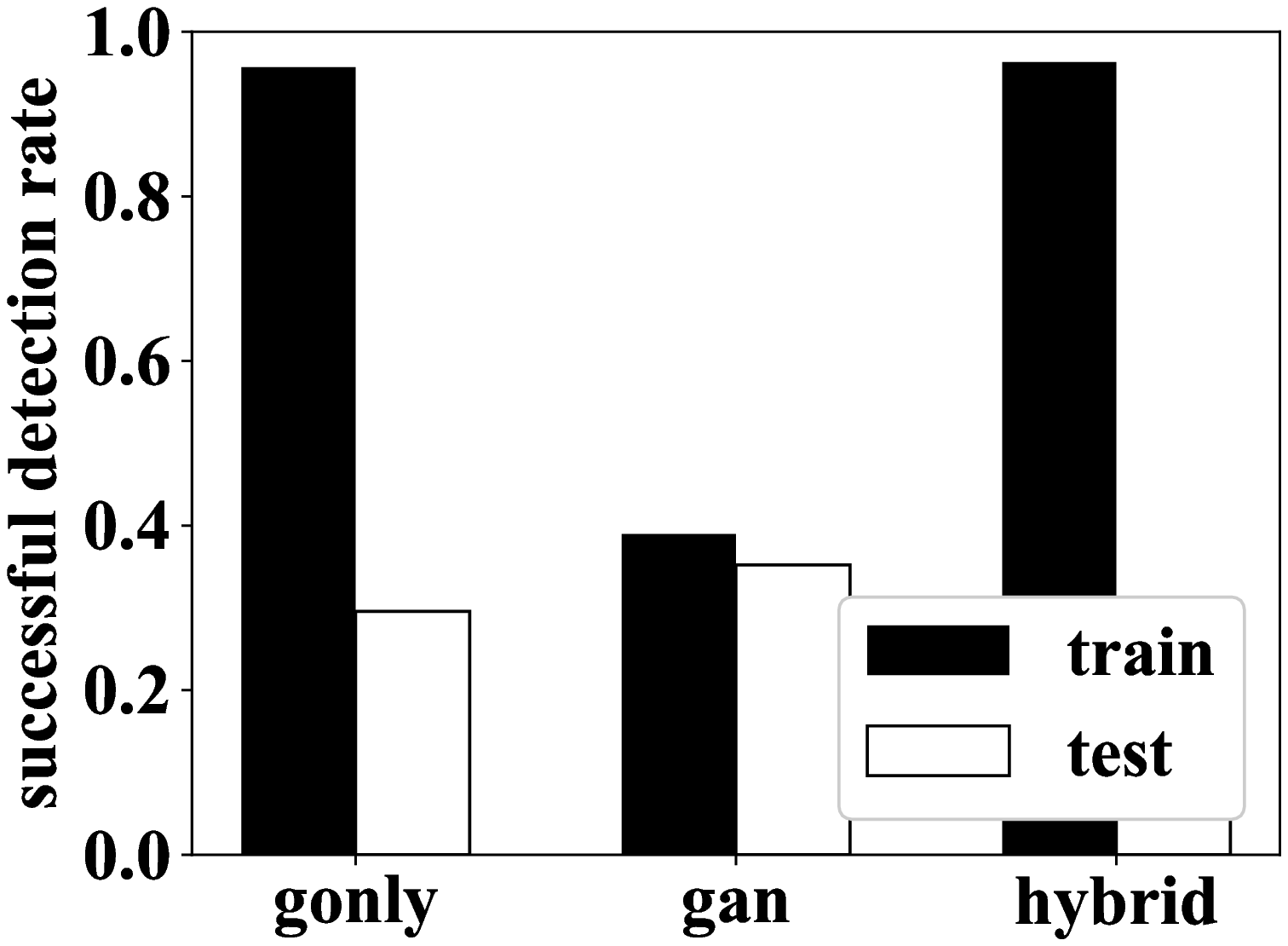}}
   \centering
   \subfloat[Average position error]{\label{fig:exp-02-01_ave_left_diffs_32000}\includegraphics[width=0.25\textwidth]{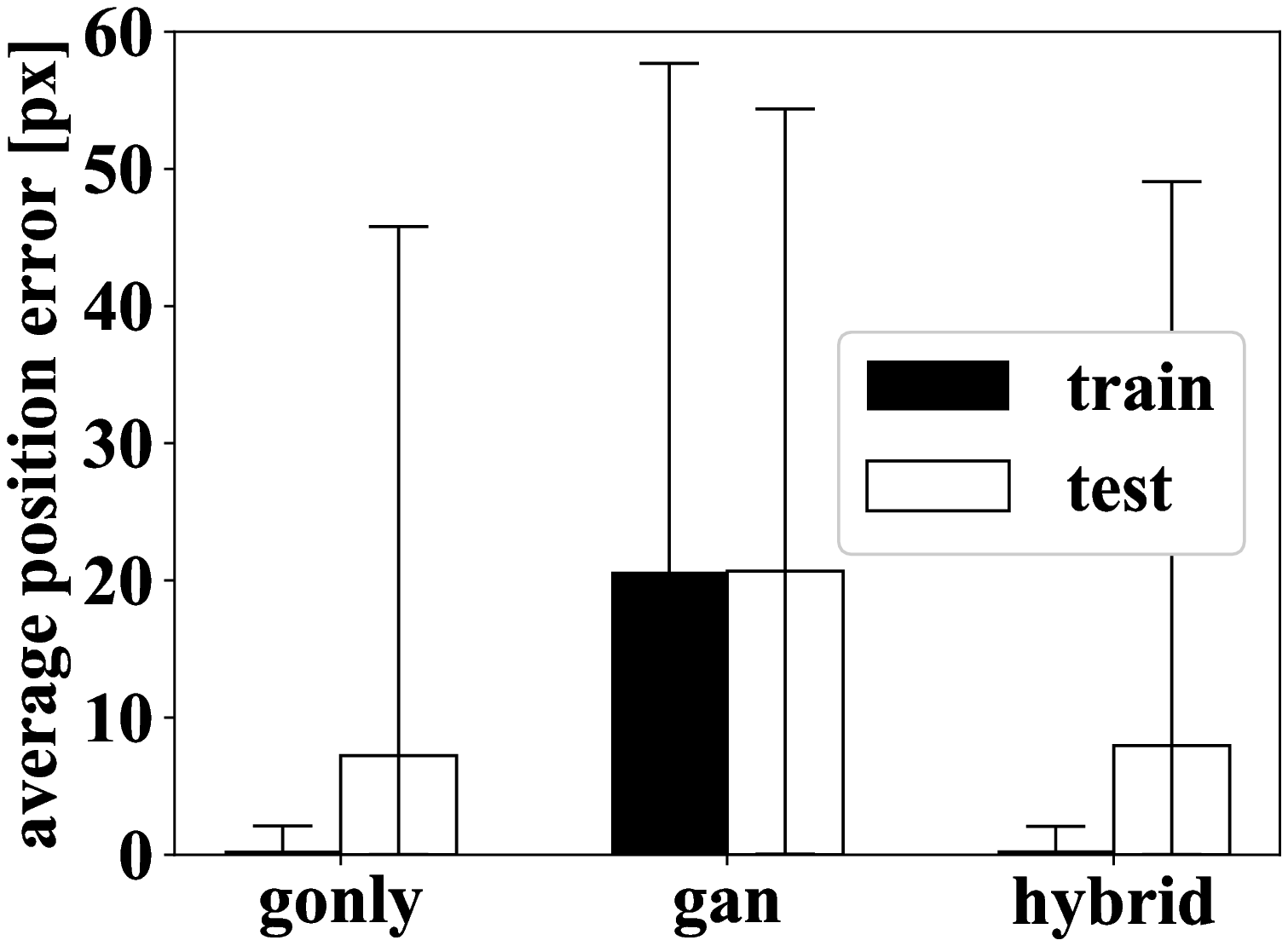}}
  \caption{Experiment 3: Single-user continuous position estimation}
  \label{fig:exp-02-01_ave_left_diffs_32000}
  \end{center}
\end{figure}

To evaluate a more complex situation than that in Section \ref{sec:eval-user-2}, experiments were conducted in which one person walked around an oval connecting positions 1 to 3, as shown in Figure ~\ref{fig:a608}.
The evaluation used 515 images as training data and 498 images as test data.
The other settings were identical to those in Experiment 1 and 2.
As the results of the qualitative evaluation did not differ from those of the position detection problem in Section~\ref{sec:eval-user-1}, only the quantitative evaluations are presented in this section.

Figure~\ref{fig:exp-02-01_ave_left_diffs_32000}\subref{fig:exp-02-01_drate_32000} shows the detection success rates of each comparison method.
The detection success rates are relatively low compared to those obtained in the evaluations in Section \ref{sec:eval-user-1} and Section \ref{sec:eval-user-2}. This was despite the fact that the results when using training data showed high detection success rates. In the training data, the values were 95.7 \% for generator-only learning, 39.0 \% for GAN-only learning, and 96.3 \% for hybrid learning, whereas in the test data, they were 29.6 \% for generator-only learning, 35.2 \% for GAN-only learning, and 27.8 \% for hybrid learning.
We believe that the amount of training data is small relative to the complexity of the problem.

Figure~\ref{fig:exp-02-01_ave_left_diffs_32000}\subref{fig:exp-02-01_drate_32000} shows the distance (in pixels) between the left coordinates of the detected box of the training data and that of the generated data. 
The lower value of the distance indicates that CSI2\-Image precisely tracks the position of a user.
This evaluation only used the generated images that successfully detect a user.
The lower limit of the error bar is the minimum value, and the upper limit is the maximum value.
The evaluation results show that GAN-only learning cannot be used for single-user continuous position detection.
While generator-only learning and hybrid learning are superior to GAN-only learning, they require performance improvements. This is because the maximum value is too high, although the results are only calculated from successfully detected images. Using the test data, the maximum differences of generator-only learning, GAN-only learning, and hybrid learning are 46 px, 54 px, and 49 px, respectively.

\section{Conclusion}

This paper proposed CSI2Image, a GAN-based CSI-to-image conversion method.
Specifically, three learning methods have been explored: generator-only learning, GAN-only learning, and hybrid learning.
An evaluation method using an image recognition algorithm has also been proposed as a quantitative evaluation method for CSI-to-image conversion.
For simple problems such as classifying the location of a person, it was found that the simplest generator-only learning model can be used.
In addition, it was observed that simple use of GANs, such as GAN-only learning, resulted in the successful generation high-quality images, but the images lacked physical space information.
Furthermore, hybrid learning, which is a combination of generator-only learning and GAN-only learning, was found to achieve superior performance under slightly more complex conditions, such as classifying the location of one or two people.
However, none of the three methods performed well in the more complex single-user continuous position detection problem.
It is concluded that further improvements can be made by redesigning the network structure, allowing the input of time-series CSI, inputting CSI from multiple devices, and converting CSI to higher-order information such as angle-of-arrival before inputting it into the DNN.

\bibliographystyle{ieeetr}
\bibliography{arxiv20_kato}
\end{document}